%% file: main.tex
\documentclass{article} 
\PassOptionsToPackage{dvipsnames,table}{xcolor}
\usepackage{iclr2026_conference}
\iclrfinalcopy

\input{math_commands.tex}

\usepackage{arxiv}

\usepackage{times}
\usepackage{xcolor}
\usepackage{url}
\usepackage{booktabs}
\usepackage{multirow}
\usepackage{graphicx}
\usepackage{amsmath}
\usepackage{microtype}
\usepackage{float}
\usepackage{titletoc}
\usepackage{wrapfig}
\usepackage[font=small,labelfont=bf]{caption}
\usepackage[most]{tcolorbox}

\IfFileExists{needspace.sty}{\usepackage{needspace}}{\newcommand{\Needspace}[1]{}}
\IfFileExists{placeins.sty}{\usepackage{placeins}}{\newcommand{\FloatBarrier}{}}

\definecolor{CiteBlue}{HTML}{1F5FA6}   
\definecolor{RQTint}{HTML}{EEF3FA}     

\usepackage[
    colorlinks=true,
    citecolor=CiteBlue,
    linkcolor=black,
    urlcolor=black,
    hypertexnames=false
]{hyperref}
\hypersetup{
    pdftitle={Learning from Runtime Feedback through Failure-Bank Self-Evolution for Vision-Language-Action Models},
    pdfauthor={Mingyue Cui, Zheyuan Liu, Yihan Zhu, Zheyuan Zhang, Meng Jiang},
}

\title{Learning from Runtime Feedback through\\ Failure-Bank Self-Evolution for\\ Vision-Language-Action Models}
\renewcommand{\shorttitle}{Failure-Bank Self-Evolution for Vision-Language-Action Models}

\author{
\textbf{Mingyue Cui}$^{*}$ \quad
\textbf{Zheyuan Liu}$^{*}$ \quad
\textbf{Yihan Zhu} \quad
\textbf{Zheyuan Zhang} \quad
\textbf{Meng Jiang} \\[3pt]
\textnormal{University of Notre Dame} \\[2pt]
\texttt{\{mcui3, zliu29\}@nd.edu}
}
\date{}

\newcommand{\method}{\textsc{FailBank}}

\newtcolorbox{rqbox}{%
  enhanced, colback=RQTint, colframe=RQTint, boxrule=0pt, arc=1.5pt,
  borderline west={2.2pt}{0pt}{CiteBlue},
  left=7pt, right=6pt, top=2.5pt, bottom=2.5pt,
  before skip=4pt, after skip=5pt}
\newcommand{\researchquestion}[2]{%
  \Needspace{3\baselineskip}%
  \begin{rqbox}
    \itshape\hypertarget{rq:#1}{\textcolor{CiteBlue}{\textbf{RQ#1}}}\quad #2
  \end{rqbox}%
}

\titlecontents{section}[1.8em]
  {\addvspace{5pt}\bfseries}
  {\contentslabel{1.8em}}{\hspace*{-1.8em}}
  {\titlerule*[0.6pc]{.}\contentspage}
\titlecontents{subsection}[4.3em]
  {\addvspace{1pt}}
  {\contentslabel{2.5em}}{\hspace*{-2.5em}}
  {\titlerule*[0.6pc]{.}\contentspage}
\titlecontents{subsubsection}[7.5em]
  {\addvspace{0pt}}
  {\contentslabel{3.2em}}{\hspace*{-3.2em}}
  {\titlerule*[0.6pc]{.}\contentspage}

\begin{document}
\raggedbottom
\maketitle
{\renewcommand{\thefootnote}{\fnsymbol{footnote}}%
\footnotetext[1]{Equal contribution.}
}
\setcounter{footnote}{0}

\begin{abstract}
Vision-language-action (VLA) models generalize broadly across robotic manipulation tasks, but complex environments require balancing task success with unintended contact. Runtime shields can correct individual actions, but they leave the underlying policy unchanged, so repeated disagreements may create a persistent policy–shield mismatch that blocks task progress. To address this challenge, we introduce \method{}, a four-stage self-evolving framework that converts runtime feedback into persistent policy improvement. During collection, a fixed CBF-based safety module serves as an observe-only teacher, producing counterfactual corrections while the policy remains in control. Outcome-aware admission then converts useful proposals into corrective targets and retains successful uncorrected actions as quiet anchors for guarded LoRA updates. We evaluate \method{} on the VLA-Arena benchmark across two difficulty levels and two VLA backbones. Compared with the base policies, \method{} improves the joint success–cost operating point. Across the two backbones, \method{} improves task success rate by 8.5 and 6.9 percentage points, while reducing policy-induced cumulative cost by 35.6\% and 23.8\%, respectively. Compared with runtime shielding, \method{} raises task success rate by 25.4 and 9.5 percentage points, while maintaining comparable policy-induced cumulative cost. These results show that runtime feedback can serve as persistent policy supervision rather than only as a temporary action constraint.%
\footnote{Code is available at {\hypersetup{urlcolor=CiteBlue}\href{https://mingyuee88.github.io/FailBank/}{Mingyuee88/FailBank}}.}
\end{abstract}

\section{Introduction}

Vision-language-action (VLA) models connect visual observations and language
instructions to continuous robot control, allowing a single policy to address
many manipulation tasks without task-specific controllers
\citep{black2024pi0,pi2025pi05}. 
However, deploying such general policies in complex scenes requires balancing
task completion with unintended contact. Real-world scenes can contain lookalike objects and nearby protected items, so the policy must identify the target, execute a precise action sequence, and avoid disturbing irrelevant objects. Task success and cumulative cost must therefore be evaluated jointly, since improving task success may increase contact, while overly conservative behavior may avoid contact at the cost of task completion.

Recent works use runtime shields, failure monitoring, and constrained learning to improve
robot safety \citep{hu2025vlsa,zhang2025safevla,gu2025safe,lyu2026foresight,english2026neurosymbolic}. A representative example of runtime shielding is AEGIS \citep{hu2025vlsa},
which  uses visual grounding and a control barrier
function (CBF) to correct unsafe actions before execution. However, this protection is only temporary, as the shield changes the robot command while the nominal policy remains fixed. The same action disagreement can therefore recur over consecutive control steps and produce repeated interventions that may push the robot into unfamiliar states.

More importantly, repeated intervention reveals a deeper limitation of runtime shielding, which can correct individual actions but cannot resolve the persistent policy--shield mismatch. In a difficult scene, such as a narrow grasp surrounded by protected objects, projection may reduce immediate cost yet
repeatedly redirect a capable policy until the episode times out. Strengthening the shielding does not remove the policy--shield mismatch because the nominal policy continues to generate the same class of action. Nevertheless, the teacher proposal provides
a candidate counterfactual target for how the policy could move under the shield's local
geometric model.

This observation leads to our central question:
\emph{Can runtime evidence be converted into learning records that enable self-evolving policy updates and improve future policy behavior?} Answering this question requires more than logging interventions, as the system must observe the policy’s failure distribution, distinguish useful pre-contact corrections from invalid actions, preserve successful behavior against drift, and prevent adapter updates from degrading the original action distribution.

To address this challenge, we propose \method{}, a four-stage self-evolving framework that uses a fixed CBF teacher during collection. In Stage~1, the policy executes nominal actions while the teacher logs counterfactual proposals. Stage~2 selects outcome-aware learning records and retains successful uncorrected actions as quiet anchors. Stage~3 accumulates the admitted records in the training bank. Finally, Stage~4 fits a fresh LoRA adapter and accepts it only if its held-out flow loss and first-action drift remain within fixed limits. The accepted policy then carries this evidence into future rollouts and requires only RGB observations and proprioception at deployment.

We evaluate \method{} on VLA-Arena's static-obstacle suite across two difficulty levels and two VLA backbones, with higher levels indicating greater task difficulty. The learned updates consistently improve the joint success–cost operating point, increasing task success while reducing policy-induced cumulative cost. These gains extend to harder tasks where shielding may sharply reduce success. 

Our contributions are
\begin{itemize}
    \setlength{\itemsep}{2pt}
    \setlength{\parskip}{0pt}
    \setlength{\parsep}{0pt}
    \setlength{\topsep}{3pt}
    \item We identify persistent policy--shield mismatch as a key limitation of action-only runtime protection and introduce an observe-only interface that collects counterfactual teacher proposals on the policy's own rollout distribution without altering execution.

    \item We propose a four-stage self-evolving framework with outcome-aware records, an accumulated failure bank, and a guarded LoRA update.
    
    \item Extensive experiments across multiple difficulty levels and VLA backbones show that \method{} improves task success rate by 8.5 and 6.9 percentage points and reduces policy-induced cumulative cost by 35.6\% and 23.8\% over the base policy, while achieving SR gains of 25.4 and 9.5 percentage points over AEGIS across the two backbones at comparable cost.
\end{itemize}
\section{Motivation}
\label{sec:motivation}

\begin{figure}[t]
  \centering
  \includegraphics[width=\textwidth]{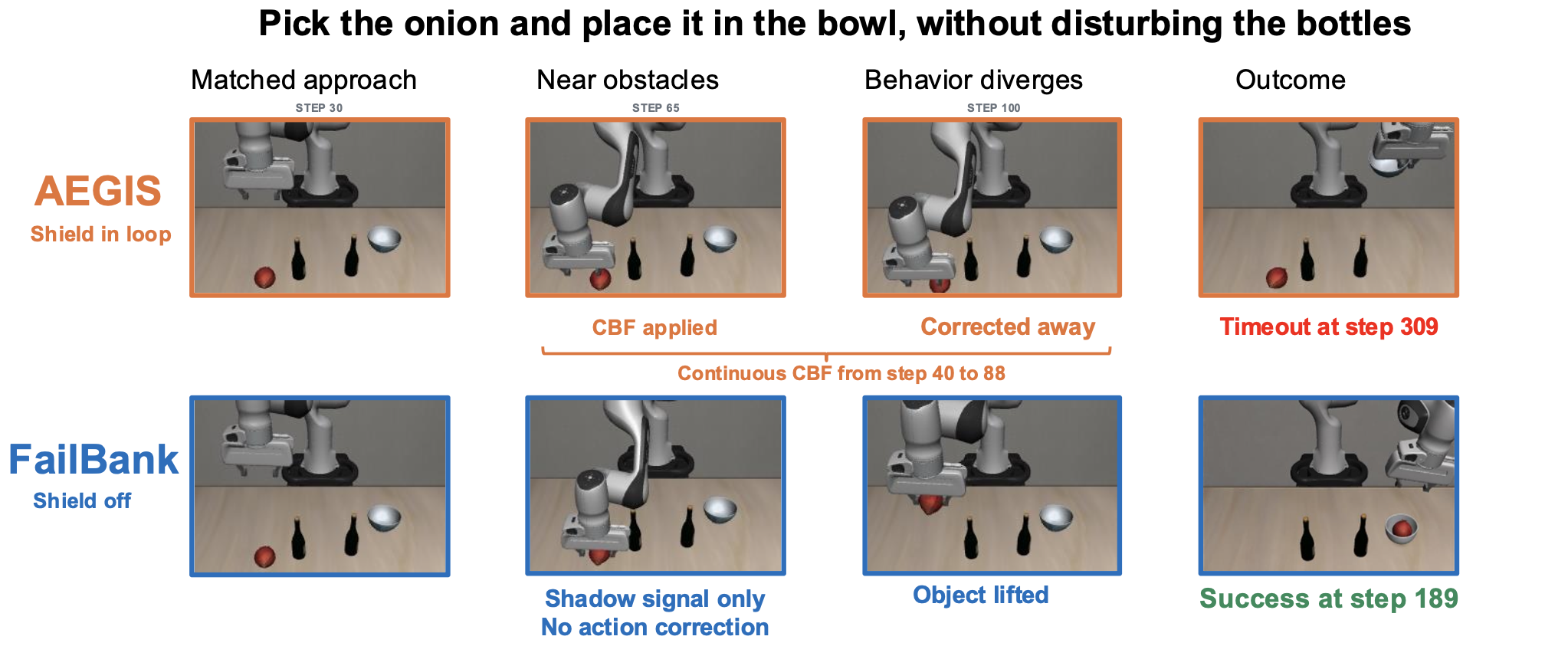}
  \caption{\textbf{A shield-collapse trajectory on the Level 2 onion task. }AEGIS repeatedly intervenes near the bottles and eventually times out. \method{} uses observe-only CBF supervision and successfully places the onion in the bowl.}
  \label{fig:qualitative}
\end{figure}

\noindent \textbf{Policy--shield mismatch.}
Runtime shielding can correct individual actions without updating the nominal policy \citep{ames2017cbf,hu2025vlsa}. When policy and shield
repeatedly mismatch, the resulting corrections may reduce local cost while
impairing task progress. Figure~\ref{fig:qualitative} illustrates this
mismatch on a harder task trajectory, where repeated shielding eventually prevents task completion. This motivates us to use runtime corrections as supervision for updating the policy, rather than applying them only during execution.

\begin{figure}[t]
  \centering
  \includegraphics[width=\textwidth]{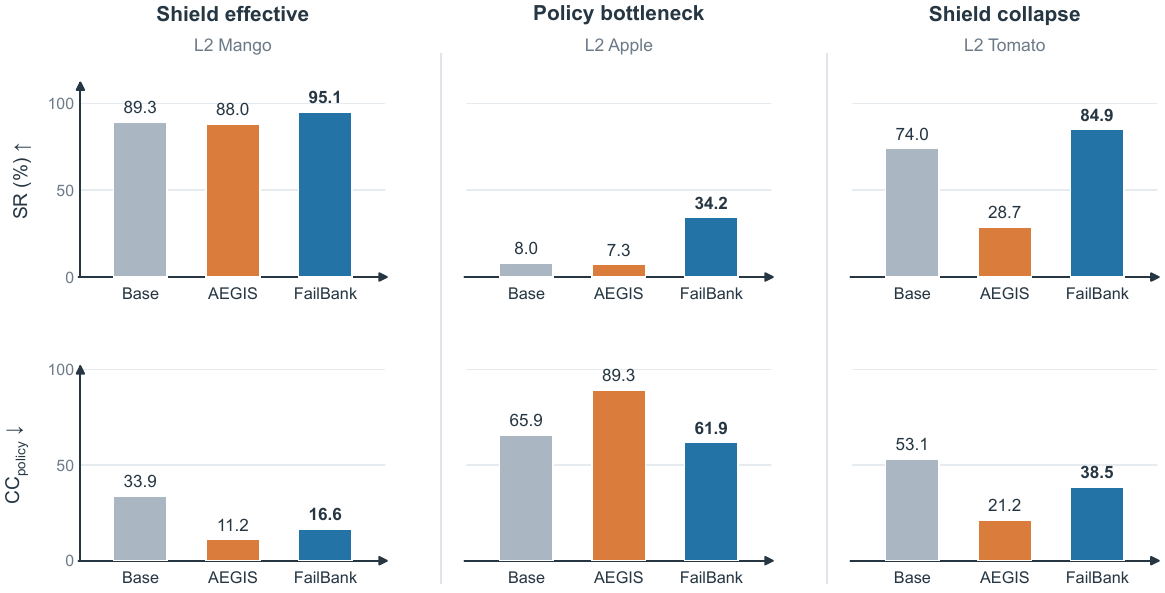}
 \caption{\textbf{Representative Level-2 static-obstacle results on VLA-Arena.}
 Top: task success rate (SR, $\uparrow$ better); bottom: policy-induced cumulative cost ($\mathrm{CC}_{\mathrm{policy}}, \downarrow$ better).
 AEGIS reduces cost while largely preserving success on Mango, fails to improve success on Apple, and sharply reduces success on Tomato.
 \method{} achieves the highest success rate across all three tasks while reducing policy-induced cost relative to Base.
}
  \label{fig:motivation}
\end{figure}

\noindent \textbf{Shield collapse.}
Runtime shielding keeps the VLA policy fixed, so repeated action disagreement can persist and eventually block task progress, especially on harder tasks. Figure~\ref{fig:motivation} illustrates this issue across three diagnostic regimes, ranging from effective shielding to a policy bottleneck and shield collapse. Across all three, \method{} maintains higher success while further reducing $\mathrm{CC}_{\mathrm{policy}}$, motivating the use of runtime corrections as policy supervision rather than only action-time protection.

\noindent \textbf{A shield-collapse trajectory.}
Figure~\ref{fig:qualitative} visualizes the policy--shield mismatch on a harder Level 2 onion task. The nominal policy continues to approach the target onion, while AEGIS repeatedly redirects the gripper away from nearby hazard bottles to reduce safety cost. Because the target lies in the same constrained region, these corrections also pull the gripper away from the onion, preventing task completion and eventually causing a timeout. \method{} instead learns from this runtime feedback and completes the same task successfully. Additional trajectory and stacking analyses are
reported in Appendix~\ref{app:coordination}.

\section{Related Work}

We provide an overview of current research on generalist VLA policies and evaluation, runtime safety mechanisms, and learning-based safety adaptation. A more detailed discussion of related work is provided in Appendix~\ref{app:related-details}.

\noindent \textbf{Vision-language-action policies and evaluation.}
Generalist VLA models combine vision-language representations with robot control.
$\pi_0$ uses flow matching for continuous action generation, while $\pi_{0.5}$
extends this family toward broader generalization
\citep{black2024pi0,pi2025pi05}. VLA-Arena evaluates such policies across
controlled safety, distractor, extrapolation, and long-horizon settings
\citep{zhang2026vlaarena}. We use its official task definitions and metrics.

\noindent \textbf{Runtime safety mechanisms.}
Control barrier functions (CBF) provide a principled mechanism for constraining nominal controls
\citep{ames2017cbf}. AEGIS combines CBF projection with visual grounding as a
plug-and-play VLA safety layer \citep{hu2025vlsa}, while constrained flow
matching incorporates safety guidance during action generation
\citep{english2026neurosymbolic}. \method{} instead uses runtime corrections as
supervision for future policy behavior.

\noindent \textbf{Learning-based safety adaptation.}
SafeVLA integrates safety through constrained learning
\citep{zhang2025safevla}, while SAFE detects failures from internal VLA
representations \citep{gu2025safe}. Privileged supervision and low-rank
adaptation provide additional foundations for transferring training-time
information into a deployable policy \citep{chen2020learning,hu2022lora}.
\method{} builds on these ideas by converting outcome-screened runtime feedback
into learning records for guarded, iterative policy updates.

\section{Method}

\subsection{Problem setting and overview}
\label{sec:method-overview}

\begin{figure}[t]
\centering
\includegraphics[width=0.96\textwidth]{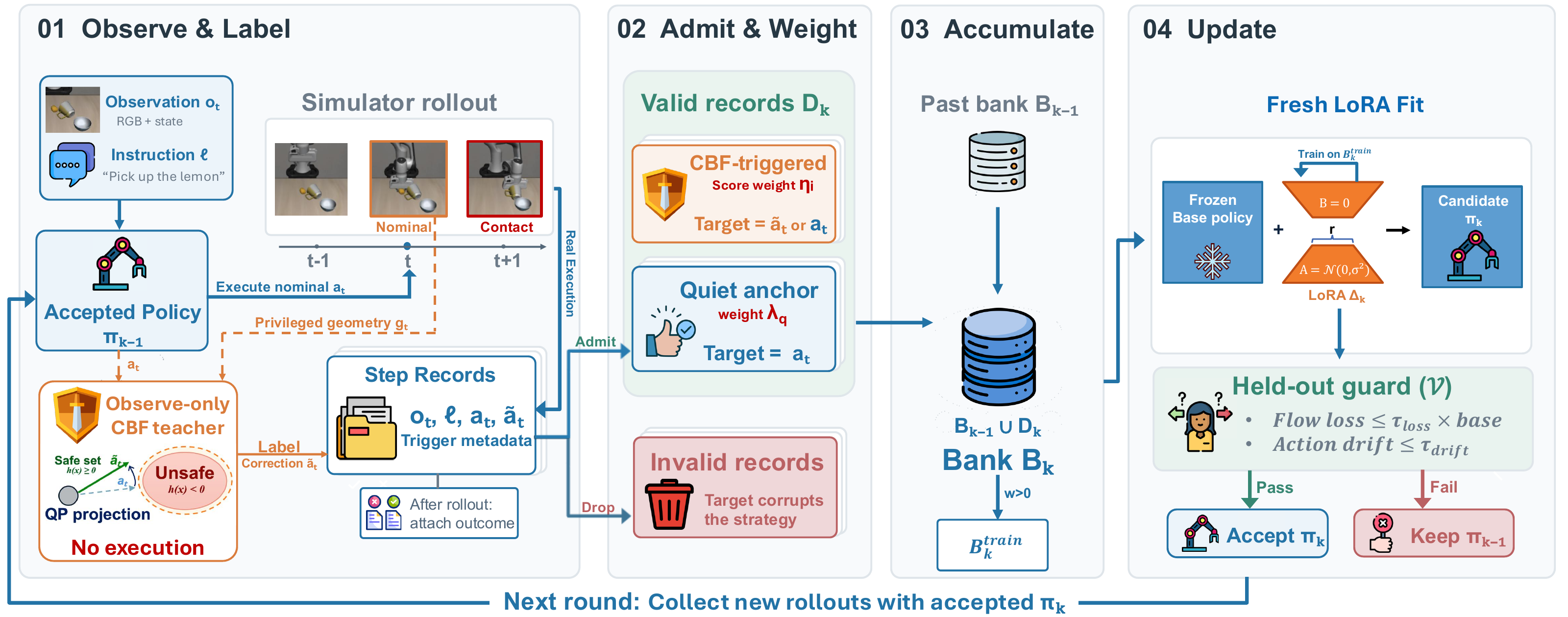}
\caption{\textbf{The four-stage \method{} loop.} Stage~1 collects policy-controlled rollouts while logging CBF proposals without altering execution. Stage~2  discards invalid records and filters valid records into CBF-triggered corrections and quiet anchors, assigning their targets and weights. Stage~3 accumulates the valid records in the training bank. Stage~4 fits a fresh LoRA adapter and accepts the candidate policy only if it passes the held-out guard. The accepted policy is then used to collect rollouts in the next round.}
\label{fig:pipeline}
\end{figure}

We denote the original policy by $\pi_{\mathrm{base}}$ and the
accepted policy collecting in round $k$ by $\pi_{k-1}$.
Unlike runtime shielding, \method{} keeps execution under the
current policy and uses proposals from a fixed CBF teacher as
counterfactual supervision. Based on rollout outcomes, it selects
corrective records and successful actions as quiet anchors, then
accumulates them in a failure bank. Each round fits a fresh LoRA
adapter from $\pi_{\mathrm{base}}$, accepting it only if it passes
a held-out guard. The accepted policy collects the next round
and is deployed without the teacher or privileged geometry. Figure~\ref{fig:pipeline} summarizes the four-stage loop from observe-only
annotation to a guarded policy update.

\subsection{Stage 1: Observe and Label}
\label{sec:observe-only}

The first stage preserves the policy's own rollout distribution while collecting
counterfactual supervision. At each control step, $\pi_{k-1}$ produces $a_t$,
and the observe-only teacher independently proposes
$\tilde a_t=S(a_t,g_t)$ using privileged scene geometry $g_t$.
The teacher changes only the translational channels and records whether
the projection was triggered through $z_t\in\{0,1\}$. The environment
always executes the nominal action,
\begin{equation}
    a_t^{\mathrm{env}}=a_t.
    \label{eq:observe-only}
\end{equation}
Because the proposal is not executed, the episode outcome remains
attributable to $\pi_{k-1}$.

Each step record stores the observation reference, instruction, nominal action, teacher proposal, trigger indicator, and metadata needed to audit the projection. After the rollout, we attach the episode outcome to these records, producing outcome-augmented evidence that Stage~2 uses to determine which records should be retained and how they should be used for learning. To isolate the learning signal from perception errors, our observe-only teacher uses privileged simulator geometry during collection. This information is not required at deployment, while the runtime-shield baseline requires geometry through visual grounding.
Appendix~\ref{app:implementation} details the interfaces, and Appendix~\ref{app:attribution} reports the matched learning-signal attribution controls.

\subsection{Stage 2: Admit and Weight}

The second stage converts rollout evidence into weighted training records $\mathcal D_k$ based on rollout outcomes and correction timing. Each admitted record is assigned a first-action target $y_i$. When a teacher correction is retained, the teacher proposal serves as the target, while successful uncorrected actions with $z_i=0$ are retained as quiet anchors and use the nominal action instead. Target assignment is independent of trigger provenance, so a CBF-triggered record may retain $z_i=1$ even when its target is the nominal
action. Such a record is not considered a quiet anchor. Records that fail the admission checks are discarded. Each admitted record is weighted by trigger provenance,
\begin{equation}
  w_i=
  \begin{cases}
    \eta_i,    & z_i=1,\\
    \lambda_q, & z_i=0.
  \end{cases}
  \label{eq:record-weight}
\end{equation}
For CBF-triggered records, $\eta_i$ is determined from the subsequent rollout: safety, task progress, and recovery increase the score, whereas repeated triggers, short-horizon cost, and barrier violations dexrease it. A validity check and fixed thresholds map the resulting score to $\eta_i\in\{0,0.25,0.60,1.00\}$. Because the teacher proposal is not executed, $\eta_i$ reflects the training utility of the record rather than the causal
effect of the correction. Quiet anchors receive the predefined weight $\lambda_q$.

Appendix~\ref{app:record-admission} gives the complete admission, lead-time, and weighting rules, while Appendix~\ref{app:recipe} reports the
$\lambda_q$ values used in our experiments.


\subsection{Stage 3: Accumulate}

Stage~3 integrates the newly admitted records $\mathcal D_k$ with evidence retained from previous rounds. To keep the held-out guard separate from training, episodes are split into training and validation folds before Stage 2 admission, with a fixed held-out batch $\mathcal V$ drawn from the validation fold and kept disjoint from $\mathcal B_k^{\mathrm{train}}$. The accumulated training bank is then updated as
\begin{equation}
  \mathcal B_k^{\mathrm{train}}
  =\mathcal B_{k-1}^{\mathrm{train}}\cup\mathcal D_k.
  \label{eq:bank}
\end{equation}
The bank carries earlier records into subsequent rounds. Bank composition and construction audits are reported in Appendix~\ref{app:bank-ledger} and Appendix~\ref{app:bank-audit}, respectively.

\subsection{Stage 4: Update}

Stage~4 turns the accumulated evidence in $\mathcal B_k^{\mathrm{train}}$ into a guarded policy update. At each round, we fit a fresh LoRA adapter from the same base checkpoint. For record $i$, let $\ell^{\mathrm{flow}}_{i,0}(\pi,y_i)$ denote the first-action flow-matching loss against its assigned target $y_i$. We optimize
\begin{equation}
  \mathcal L(\pi)
  =\frac{
    \sum_{i\in\mathcal B_k^{\mathrm{train}}}
    w_i\ell^{\mathrm{flow}}_{i,0}(\pi,y_i)}{
    \max\!\left(1,\sum_{i\in\mathcal B_k^{\mathrm{train}}}w_i\right)}.
  \label{eq:objective}
\end{equation}
The resulting candidate is then evaluated on the fixed held-out batch $\mathcal V$ using full-chunk flow loss and first-action drift
from the original base policy,
\begin{equation}
  \bar{\mathcal L}_{\mathcal V}(\pi)
  =\frac{1}{|\mathcal V|H}
    \sum_{i\in\mathcal V}\sum_{h=0}^{H-1}
    \ell^{\mathrm{flow}}_{i,h}(\pi),
  \qquad
  D_{\mathcal V}(\pi)
  =\frac{1}{|\mathcal V|d}
    \sum_{i\in\mathcal V}
    \left\lVert a^{(\pi)}_{i,0}
    -a^{(\mathrm{base})}_{i,0}\right\rVert_1.
  \label{eq:guard-metrics}
\end{equation}
The candidate is accepted only if
\begin{equation}
  \frac{\bar{\mathcal L}_{\mathcal V}(\pi_k)}
       {\bar{\mathcal L}_{\mathcal V}(\pi_{\mathrm{base}})}
  \leq\tau_{\mathrm{loss}},
  \qquad
  D_{\mathcal V}(\pi_k)\leq\tau_{\mathrm{drift}}.
  \label{eq:guard}
\end{equation}

These constraints limit held-out loss degradation and first-action drift
without using benchmark SR or CC for model selection. If the candidate fails,
the adapter is discarded and $\pi_{k-1}$ remains the collecting policy, while
$\mathcal B_k^{\mathrm{train}}$ is retained. If it passes, the accepted
$\pi_k$ becomes the collecting policy for the next round and is deployed
without the teacher or privileged geometry.

Appendix~\ref{app:recipe} gives the training recipe, while
Appendix~\ref{app:guard-diagnostics} reports the guard limits and diagnostics.

\section{Experiments}
\label{sec:experiments}


We evaluate \method{} through four research questions:
\textbf{(1)} Can runtime feedback improve task success while reducing policy-induced cost? \textbf{(2)} Does the learned policy generalize across states, tasks, levels, and backbones? \textbf{(3)} How do observe-only collection and learning signals affect policy updates?
\textbf{(4)} How does iterative self-evolution affect the success--cost operating point?

\subsection{Experimental setup}
\label{sec:setup}

\noindent \textbf{Benchmark and task coverage.}
VLA-Arena organizes manipulation tasks into Safety, Distractor, Extrapolation,
and Long-Horizon categories \citep{zhang2026vlaarena}. We evaluate its
static-obstacle safety suite, with five tasks at each of three difficulty levels.
The released Arena checkpoints are finetuned on Level~0 demonstrations, so we use
Levels~1 and~2 to study improvement beyond that source difficulty. We collect on
Level~1 mango, test all five Level~1 tasks, and test all five harder Level~2 tasks
without Level~2 update data. Appendix~\ref{app:task-coverage} lists the tasks, objects, splits, and coverage.

\noindent \textbf{Evaluation metrics.}
We report success rate (SR), cumulative cost (CC), policy-induced cumulative cost ($\mathrm{CC}_{\mathrm{policy}}$) and the base-relative score (BRS). SR is the percentage of trials that satisfy the task-completion predicate within the episode limit. CC is VLA-Arena's official benchmark metric, computed as the trial-average sum of per-step costs. We additionally report $\mathrm{CC}_{\mathrm{policy}}$, which removes the cost already present in the initial state from official CC. To compare joint
improvements in success and cost,
we define the base-relative score (BRS) as
\begin{equation}
 \mathrm{BRS}=\exp\!\left[-\frac{1}{2}\left(
 \frac{1-\mathrm{SR}}{1-\mathrm{SR}_{\mathrm{base}}}
 +\frac{\mathrm{CC}_{\mathrm{policy}}}{\mathrm{CC}_{\mathrm{policy},\mathrm{base}}}
 \right)\right]
 \label{eq:score-b}
\end{equation}
where SR is expressed as a fraction. BRS equally weights the
failure rate and policy-induced cost, normalized by their
respective task-specific base values. The base policy scores
$e^{-1}$, while a policy with zero failure and zero
policy-induced cost scores 1. Appendix~\ref{app:metrics} provides the
exact calculations.

\noindent \textbf{VLA backbones.}
Our main experiments use the Arena-finetuned $\pi_{0.5}$ checkpoint, with
$\pi_0$ providing a second flow-matching backbone for cross-backbone evaluation
\citep{black2024pi0,pi2025pi05}. We audited all 29 models on the Arena
leaderboard, of which 10 provide Arena-finetuned weights. Among these models,
only $\pi_{0.5}$ and $\pi_0$ combine a continuous flow-matching action head with
measurable baseline headroom, both of which are required by our update and guard.
Appendix~\ref{app:backbone-scope} documents this selection and the attempted
extensions.

\noindent \textbf{Baselines.}
We use the Arena-finetuned $\pi_{0.5}$ flow-matching VLA as our main base policy.
We compare against the base policy and AEGIS, a CBF-based runtime
shield with GLM-4.5V perception, to distinguish persistent policy adaptation from
runtime action correction. We further repeat the Base--AEGIS--\method{} comparison
on $\pi_0$ for cross-backbone evaluation. All methods are evaluated under matched
initial states, task definitions, and evaluation conditions.
\Needspace{4\baselineskip}
\subsection{Main results}
\label{sec:main-results}
\researchquestion{1}{Can runtime feedback improve task success while reducing policy-induced cost? }

\begin{figure}[t]
\centering
\includegraphics[width=0.6\linewidth]{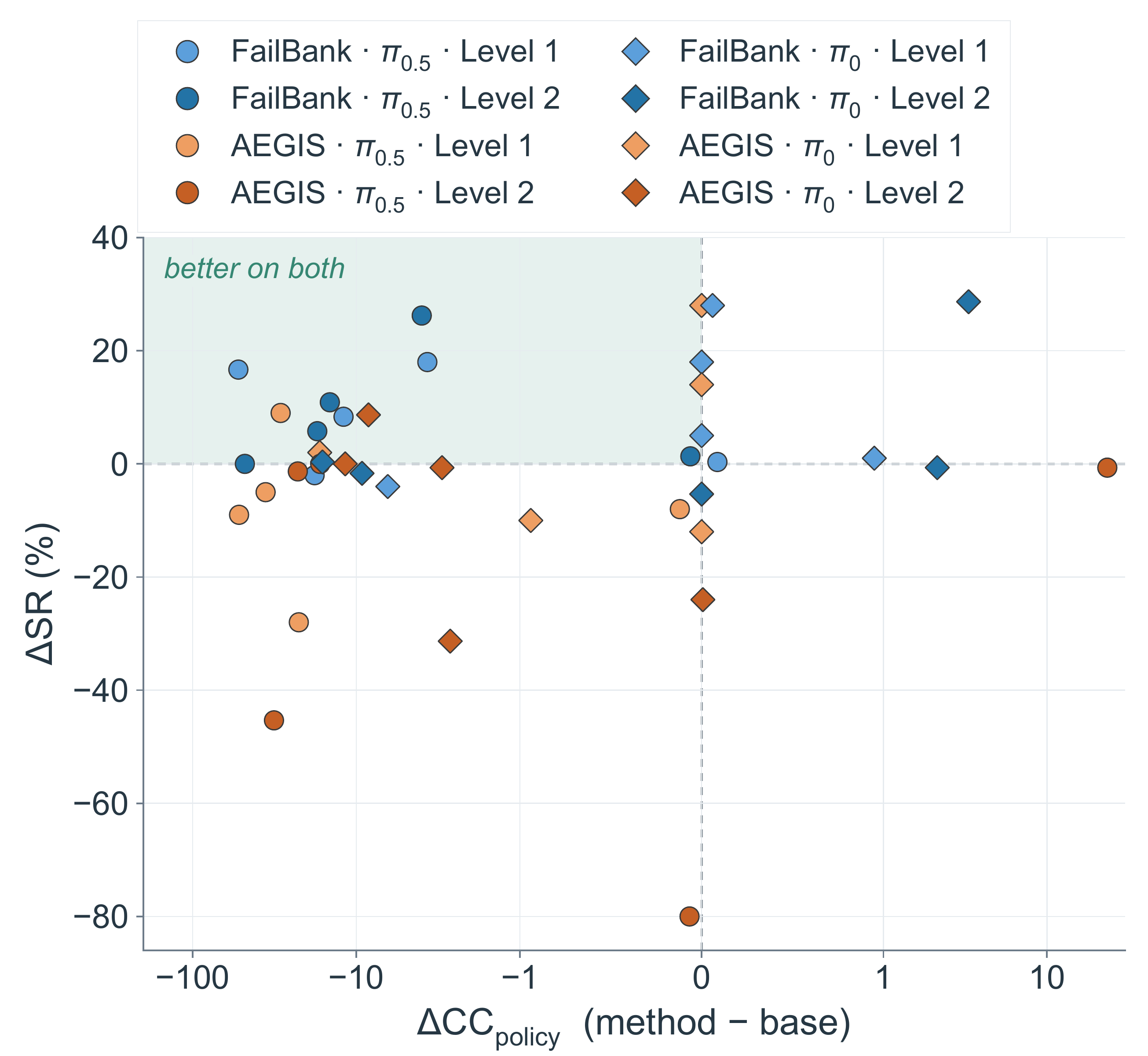}
\caption{\textbf{Success--cost trade-off relative to the base policy.}
Circles and diamonds denote \(\pi_{0.5}\) and \(\pi_0\), blue and orange denote \method{} and AEGIS, and shade indicates difficulty level. Upward movement means higher success, and leftward movement means lower policy-induced cost. Thus, the shaded upper-left quadrant is better on both. \method{} places more points in this joint-improvement region, while AEGIS more often moves left but downward.
}
\label{fig:sr-cc-tradeoff}
\end{figure}

To answer RQ1, we compare Base, AEGIS, and \method{} on the complete
Level 1 and Level 2 static-obstacle suites across both backbones.
Table~\ref{tab:summary} reports the task-level SR and cost results, while
Figure~\ref{fig:sr-cc-tradeoff} visualizes the change of each method relative to its corresponding base policy.

\input{figures/summary_table_two_backbones}
Compared with the base policies, \method{} improves both SR and
$\mathrm{CC}_{\mathrm{policy}}$ on both backbones. As shown in
Table~\ref{tab:summary}, the unweighted mean SR across the ten tasks
increases from 66.0\% to 74.5\% on $\pi_{0.5}$ and from 47.5\%
to 54.4\% on $\pi_0$, corresponding to gains of approximately
8.5 and 6.9 percentage points, respectively. Mean
$\mathrm{CC}_{\mathrm{policy}}$ decreases from 47.76 to 30.76
on $\pi_{0.5}$ and from 10.62 to 8.09 on $\pi_0$, giving
relative reductions of 35.6\% and 23.8\% over the base policies. Compared with AEGIS runtime shielding, \method{} increases mean SR
from 49.1\% to 74.5\% on $\pi_{0.5}$ and from 44.9\% to 54.4\%
on $\pi_0$, yielding gains of 25.4 and 9.5 percentage points,
respectively. Across the ten tasks, \method{} achieves mean $\mathrm{CC}_{\mathrm{policy}}$ of 30.76 on $\pi_{0.5}$\
and 8.09 on $\pi_0$, compared with 29.01 and 6.27\
for AEGIS, respectively. Thus,
\method{} improves both success and cost relative to the base
policies, while its advantage over AEGIS is higher task success
at an additional policy-induced cost.

Figure~\ref{fig:sr-cc-tradeoff} makes this joint improvement more explicit.
Under the criterion of higher SR and lower
$\mathrm{CC}_{\mathrm{policy}}$ than base policy,
\method{} achieves joint improvements in SR and
$\mathrm{CC}_{\mathrm{policy}}$ on 8 task--backbone pairs,
compared with 3 for AEGIS. AEGIS more often moves
left toward lower cost but also downward toward lower success, particularly on
harder tasks. In contrast, \method{} improves both
objectives on a larger share of tasks. This pattern is also reflected in aggregate BRS, where \method{} scores higher than AEGIS on both backbones.

\Needspace{4\baselineskip}
\subsection{Generalization across states, tasks, levels, and backbones}
\label{sec:transfer}
\researchquestion{2}{Does the learned policy generalize across states, tasks, levels, and backbones?}

To answer RQ2, we test whether the learned behavior extends beyond the
Level~1 mango collection data. We consider held-out states, unseen tasks,
and the harder Level~2 setting, and repeat the update on $\pi_0$ to test
a second backbone. Table~\ref{tab:summary} reports results across both
backbones and difficulty levels, while Appendix~\ref{app:generalization-radii}
details the evaluation splits.

On held-out initial states of the Level~1 mango task, \method{} raises
$\pi_{0.5}$ SR from $71.1\%$ to $93.3\%$, showing improvement beyond the
collection states. Across the four unseen Level~1 tasks, mean SR increases
from $82.5\%$ to $88.7\%$, extending the gains beyond the collection task.
The improvements also carry over to the harder Level~2 setting, where
mean SR rises from $50.5\%$ to $59.4\%$ without any Level~2 rollouts
entering the main update bank.

To test a second backbone, we apply the same update procedure to $\pi_0$.
Its mean SR increases from $51.0\%$ to $62.7\%$ across the four unseen
Level~1 tasks and from $35.7\%$ to $40.0\%$ on Level~2.
Together, these results provide evidence of transfer across states, tasks,
difficulty levels, and VLA backbones.
Complete paired tests and task-specific results are reported in
Appendix~\ref{app:full-stats}.

\section{Discussion}
\label{sec:discussion}

 We next examine the contribution of each learning component and how
iterative self-evolution affects policy updating performance.

\subsection{Collection interface and learning signals}
\label{sec:learning-signals}
\researchquestion{3}{How do observe-only collection and learning signals
affect policy updates?}

To answer RQ3, we remove the key collection and learning components of
\method{} in turn and examine how each affects the resulting update.

\begin{figure}[t]
  \centering
  \includegraphics[width=0.92\textwidth]{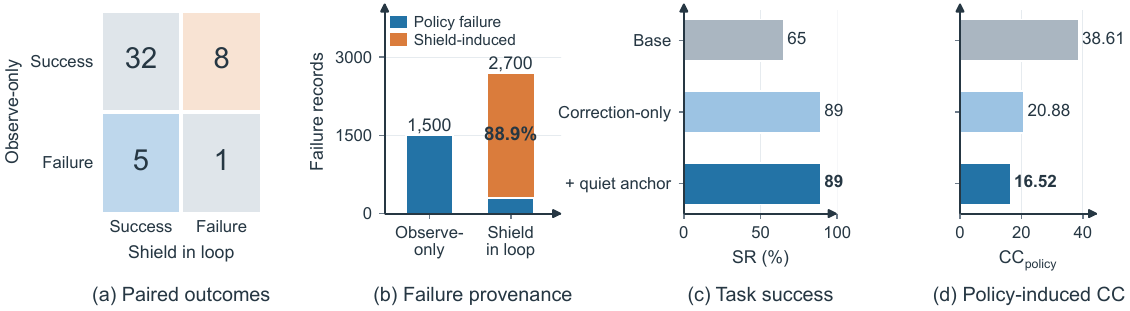}

\caption{\textbf{Ablations of the collection interface and learning signals.}
(a) Paired episode outcomes under observe-only and shield-in-loop
collection on Level 1 mango task. Rows show observe-only outcomes, while columns show
outcomes when shield corrections are executed.
(b) Failure records produced on the same task by the two collection modes, separated by policy
failures and shield-induced failures. (c--d)  Comparison of Base, training on CBF-triggered records
(Correction-only), and training with additional successful
uncorrected actions (+ quiet anchors), reporting SR and
$\mathrm{CC}_{\mathrm{policy}}$ on Level 1 onion task, respectively.}
  \label{fig:learning-signal-ablation}
\end{figure}

\noindent\textbf{Stage 1 ablation: Shield-in-loop collection.}
\label{sec:collection-mode}
Executing the shield changes the rollout distribution from which learning records are collected. As shown in Figure~\ref{fig:learning-signal-ablation}(a),
executing the shield turns eight otherwise successful policy
rollouts into failures while rescuing only five failures,
reducing successful episodes from 40 to 37 out of 46.
More importantly, Panel~(b) shows that 88.9\% of the
failure records collected with the shield in the loop
originate from these shield-induced failures rather than
failures of the nominal policy. These records therefore reflect failures induced by the collection process rather
than failures of the policy on its original rollout distribution. Observe-only collection avoids this shift and better preserves the failure distribution of the
policy being updated.

\noindent\textbf{Stage 2 ablation: Removing quiet anchors.}
Quiet anchors preserve successful policy actions that require no CBF correction, helping prevent the update from overfitting to corrective records and drifting
away from already effective behavior. Panels~(c) and~(d) of Figure~\ref{fig:learning-signal-ablation} isolate this effect by comparing an update trained only on CBF-triggered records with one that additionally includes
quiet anchors. The correction-only update raises SR from $65\%$ to $89\%$ and reduces $\mathrm{CC}_{\mathrm{policy}}$ from $38.61$ to $20.88$. Adding quiet anchors preserves the same $89\%$ SR while further reducing $\mathrm{CC}_{\mathrm{policy}}$ to $16.52$. These results suggest that corrective
records drive most of the task-success improvement, while quiet anchors help preserve successful behavior and further reduce safety cost. Further analyses of collection outcomes, failure-record provenance, and learning-signal controls are provided in Appendices \ref{app:collection-outcomes},
\ref{app:collection-provenance}, and~\ref{app:attribution}.

\Needspace{22\baselineskip}
\subsection{Number of Accumulated Self-Evolution Rounds}

\researchquestion{4}{How does iterative self-evolution affect the success--cost operating point?}
\vspace{-4pt}

\begin{wrapfigure}{r}{0.49\textwidth}
    \vspace{-10pt}
    \centering
    \includegraphics[width=0.97\linewidth]{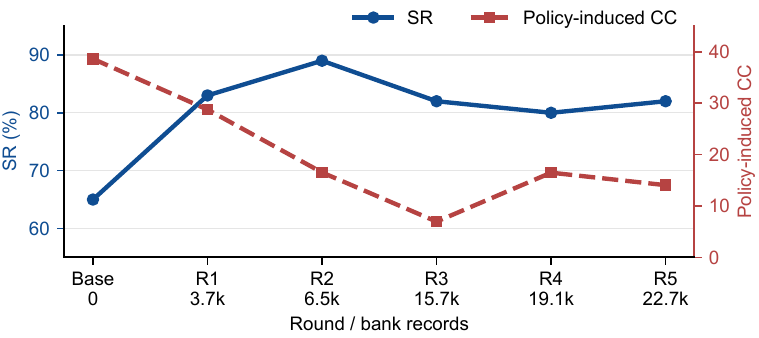}
    \caption{\textbf{Self-evolution across accumulated rounds on Level 1 onion task.}
    The failure bank grows from $3.7$k records at R1 to $22.7$k at R5.
    SR peaks at R2, whereas $\mathrm{CC}_{\mathrm{policy}}$ reaches its
    minimum at R3. Later rounds fluctuate while remaining improved over
    base policy on both axes.}
    \label{fig:rounds}
    \vspace{-8pt}
\end{wrapfigure}

We study five accumulated self-evolution rounds on the static-obstacle Level 1 onion task. As the accumulated failure bank grows from $3.7$k records at Round~1 to
$22.7$k at Round~5, Figure~\ref{fig:rounds} shows a clear evolution of the
SR--$\mathrm{CC}_{\mathrm{policy}}$ operating point. The first two rounds improve both objectives relative to Base.
Specifically, SR peaks at Round~2, while
$\mathrm{CC}_{\mathrm{policy}}$ continues to decrease
and reaches its minimum at Round~3.
The two objectives therefore reach their best values
at different rounds.
In later rounds, SR decreases and then partially recovers, while
policy-induced CC rebounds from its Round~3 minimum and continues to fluctuate.
Thus, additional runtime feedback continues to reshape the learned policy
rather than monotonically improving either objective. Importantly, all five
rounds remain above Base in SR and below Base in policy-induced CC, indicating
that the learned improvement persists as the balance between the two objectives
changes. 

\subsection{Limitations}
\label{sec:limitations}

Our current implementation targets continuous flow-matching policies, while other action formulations require adapted supervision and guard objectives. We also focus on static obstacles, since dynamic scenes additionally require temporal obstacle prediction and teacher corrections for moving hazards. These extensions concern the form of the teacher and update interface, rather than our central focus on converting runtime feedback into persistent policy improvement.
\section{Conclusion}

\method{} turns observe-only runtime feedback into outcome-aware learning records for persistent policy improvement. By accumulating these records across rounds and applying guarded LoRA updates, the framework transfers runtime corrections into the underlying policy without requiring a shield at deployment. Across two flow-matching backbones, \method{} improves the joint success--cost operating point and generalizes beyond the collection setting. These results suggest that runtime safety feedback can serve not only as a temporary intervention mechanism, but also as supervision for improving future policy behavior.
\begingroup
\small
\bibliography{references}
\bibliographystyle{iclr2026_conference}
\endgroup

\appendix


\clearpage
\section*{Appendix Outline}
\startcontents[appendix]
\begingroup
\hypersetup{linkcolor=black}
\setlength{\parskip}{0pt}
\printcontents[appendix]{}{1}{\setcounter{tocdepth}{3}}
\endgroup
\makeatletter
\setlength{\@fptop}{0pt}
\makeatother


\section{Detailed Related Work}
\label{app:related-details}

\subsection{Generalist Vision-Language-Action Policies}
\label{app:related-vla}

Scalable robot-policy pretraining began to connect large, heterogeneous robot
datasets with transformer policies. RT-1 demonstrated real-world control at
scale, while RT-2 transferred visual-language knowledge into robotic actions
\citep{brohan2023rt1roboticstransformerrealworld,brohan2023rt2visionlanguageactionmodelstransfer}.
Open X-Embodiment broadened this direction through cross-embodiment data and
RT-X models \citep{embodimentcollaboration2025openxembodimentroboticlearning}.
Octo and OpenVLA further provided open generalist policies for manipulation
\citep{octomodelteam2024octoopensourcegeneralistrobot,kim2024openvlaopensourcevisionlanguageactionmodel}.
The broader design space includes efficient state-space architectures, compact
policies, cognition--action decoupling, tokenized actions, and latent actions
\citep{liu2024robomambaefficientvisionlanguageactionmodel,
wen2025tinyvlafastdataefficientvisionlanguageaction,
li2024cogactfoundationalvisionlanguageactionmodel,
pertsch2025fastefficientactiontokenization,
shukor2025smolvlavisionlanguageactionmodelaffordable,
bu2025univlalearningacttaskcentric}. Diffusion Policy and Action Chunking with
Transformers provide related precedents for continuous generative control and
chunked imitation learning
\citep{chi2024diffusionpolicyvisuomotorpolicy,zhao2023learningfinegrainedbimanualmanipulation}.

$\pi_0$ uses flow matching for continuous action generation
\citep{black2024pi0}, while $\pi_{0.5}$ extends this family toward broader
open-world generalization \citep{pi2025pi05}. These models provide the
flow-matching policy backbones studied in our experiments. \method{} does not
modify their action-generation architecture. It studies how runtime evidence becomes persistent supervision for the underlying policy.

\subsection{Runtime Shields and Constrained Action Generation}
\label{app:related-shields}

Control barrier functions, abbreviated as CBFs, define safety constraints over system states and
commonly use a quadratic program to project a nominal control onto a
constraint-satisfying action \citep{ames2017cbf}. AEGIS applies this pattern to
VLA control by grounding protected objects and applying CBF-based projection
before execution \citep{hu2025vlsa}. Neuro-symbolic safety guidance instead
incorporates constraints directly into flow-matching action generation
\citep{english2026neurosymbolic}. Beyond VLA control, constrained policy
optimization incorporates constraints into policy learning, and Recovery RL
separates task behavior from a learned recovery policy
\citep{achiam2017constrainedpolicyoptimization,thananjeyan2021recoveryrlsafereinforcement}.
Safety layers have also been studied for robotic manipulation in human
environments and for dynamic high-dimensional robot tasks
\citep{thumm2022provablysafedeepreinforcement,liu2023safereinforcementlearningdynamic}.
These approaches establish constraint enforcement during either policy learning
or action execution.

\method{} uses the corrective signal differently. The CBF projection acts only
as an observe-only teacher during collection. The nominal policy controls the
rollout, while the counterfactual correction is recorded but not executed. The
retained feedback is then used to update the policy for subsequent rollouts and
shield-free deployment.

\subsection{Safety Alignment and Failure Monitoring}
\label{app:related-learning}

A complementary line of work incorporates safety into policy learning or detects
unsafe behavior during execution. SafeVLA combines risk elicitation with
constrained reinforcement learning to align task behavior and safety objectives
\citep{zhang2025safevla}. SAFE learns multitask failure detectors from internal
VLA representations and supports runtime responses to predicted failures
\citep{gu2025safe}. Model-based runtime monitoring has also been coupled with
interactive imitation learning so that execution-time signals inform subsequent
policy improvement \citep{liu2023modelbasedruntimemonitoringinteractive}.

\method{} learns from corrections already produced by a safety teacher. Teacher proposals and rollout outcomes form explicit learning records. Successful uncorrected actions provide quiet anchors, and the adapter guard bounds held-out flow-loss degradation and first-action drift.

\subsection{Safety Benchmarks and Evaluation}
\label{app:related-benchmarks}

Robot-learning benchmarks measure complementary aspects of generalization.
CALVIN evaluates language-conditioned long-horizon manipulation, LIBERO studies
knowledge transfer in lifelong learning, SimplerEnv evaluates real-world robot
policies in simulation, and RoboCasa provides large-scale simulation of everyday
tasks \citep{mees2022calvinbenchmarklanguageconditionedpolicy,
liu2023liberobenchmarkingknowledgetransfer,
li2024evaluatingrealworldrobotmanipulation,
nasiriany2024robocasalargescalesimulationeveryday}. VLA-Arena adds controlled
Safety, Distractor, Extrapolation, and Long-Horizon categories with hierarchical
difficulty levels and official success and cumulative-cost metrics
\citep{zhang2026vlaarena}. ForesightSafety-VLA complements endpoint metrics with
a diagnostic taxonomy of safety failures across the VLA pipeline
\citep{lyu2026foresight}. Together, these efforts motivate evaluating task
completion together with safety-related behavior rather than success alone.

Our experiments retain VLA-Arena's official task definitions, success predicates,
and CC aggregation. We report policy-induced CC as a labelled
diagnostic decomposition of benchmark CC because some Arena instances contain
cost already present in the recorded initial state. Appendix~\ref{app:metrics}
provides the exact definitions.

\subsection{Runtime Feedback as Policy Supervision}
\label{app:related-feedback}

DAgger established the principle of collecting corrective labels on the
learner's own state distribution \citep{ross2011reductionimitationlearningstructured}.
Interactive imitation learning extends this idea through intermittent expert
feedback, including intervention-based HG-DAgger and remote-teleoperation
correction \citep{celemin2022interactiveimitationlearningrobotics,
kelly2019hgdaggerinteractiveimitationlearning,
mandlekar2020humanintheloopimitationlearningusing}. Human--AI copilot methods
similarly use intervention to improve a task policy, while residual policy
learning represents corrections as a learned addition to an existing controller
\citep{li2022efficientlearningsafedriving,silver2019residualpolicylearning}.
Privileged learning allows a teacher to use information unavailable to the
deployed student \citep{chen2020learning}. In our setting, privileged simulator
geometry is available to the collection-time teacher, while the deployed policy
receives only RGB observations and proprioception. Low-rank adaptation provides
a parameter-efficient mechanism for updating the frozen VLA backbone
\citep{hu2022lora}.

These established components form the four-stage \method{} loop: observe and label, admit and weight, accumulate, and update. Policy-controlled rollouts expose the
current behavior distribution. Outcome-aware selection converts counterfactual
corrections into provenance-bearing learning records. The failure bank retains
records across collection rounds, and a held-out guard determines whether a
candidate policy update is accepted. The loop uses runtime safety feedback to supervise future policy behavior.

\section{VLA-Arena Scope and Task Coverage}
\label{app:benchmark}

\subsection{Benchmark Hierarchy}
\label{app:benchmark-hierarchy}

VLA-Arena contains 170 tasks in 11 suites spanning Safety, Distractor,
Extrapolation, and Long-Horizon categories. Each of the five Safety suites has
five tasks at each of three hierarchical levels. L0 contains basic tasks with
clear objectives, L1 introduces intermediate complexity, and L2 contains the
most challenging scenarios. We use the official task definitions, success
conditions, and CC aggregation without modifying their semantics.

\subsection{Evaluation Metrics}
\label{app:metrics}

The equations below define SR, official CC, and its policy-induced decomposition. BRS uses Equation~\ref{eq:score-b}, with SR expressed as a fraction. Appendix~\ref{app:score-b} explains its normalization, undefined-denominator cases, and uncertainty estimates.

Let $N$ be the number of evaluation trials, $s_n\in\{0,1\}$ the benchmark
success indicator for trial $n$, and $T_n$ its number of executed control steps, capped at 300. Let $c_{n,t}$ be VLA-Arena's benchmark CC at step $t$, and let
$c_n^{\mathrm{init}}$ be the cost attributed to the recorded initial state. We
compute
\begin{equation}
  \mathrm{SR}=\frac{100}{N}\sum_{n=1}^{N}s_n,\qquad
  \mathrm{CC}
  =\frac{1}{N}\sum_{n=1}^{N}\sum_{t=1}^{T_n}c_{n,t},
  \label{eq:metrics-official}
\end{equation}
and the diagnostic decomposition
\begin{equation}
  \begin{aligned}
  \mathrm{CC}_{\mathrm{policy}}
  &=\frac{1}{N}\sum_{n=1}^{N}
  \left(\sum_{t=1}^{T_n}c_{n,t}-c_n^{\mathrm{init}}\right),\\
  \mathrm{CC}_{\mathrm{init}}
  &=\frac{1}{N}\sum_{n=1}^{N}c_n^{\mathrm{init}},\\
  \mathrm{CC}
  &=\mathrm{CC}_{\mathrm{init}}+\mathrm{CC}_{\mathrm{policy}}.
  \end{aligned}
  \label{eq:metrics-policy}
\end{equation}
Failures, timeouts, and zero-cost trials all remain in the denominator $N$.
Reported task-level values are arithmetic means over the stated matched initial
states. When a protocol repeats sampler-advance conditions for the same initial
state, those repeats are first averaged within that state before a paired test.
Policy-induced CC is labelled as a diagnostic decomposition and never replaces
the official benchmark metric.

\subsection{Complete Per-Task Results}
\label{app:complete-results}

Table~\ref{tab:main} reports all five static-obstacle tasks at both levels for each backbone. Official SR and CC are shown alongside the diagnostic policy-induced cost and BRS. Guard-rejected training orders are excluded from the reported \method{} mean, as specified in the caption.

\begin{table}[htbp]
\centering
\caption{\textbf{VLA-Arena static-obstacle results across two backbones.}
Cost cells report official $\mathrm{CC}$ followed by diagnostic $\mathrm{CC}_{\mathrm{policy}}$. BRS summarizes base-relative failure and policy-induced cost. Dashes denote undefined BRS when the base policy-induced cost is zero.
\method{} averages three training orders on $\pi_{0.5}$ and the two of three that pass the training guard on $\pi_0$.
L1-T2 is the collection task. L2-T1 is a floor task on both backbones, and so is L2-T0 on $\pi_0$.}
\label{tab:main}
\scriptsize
\setlength{\tabcolsep}{2.2pt}
\resizebox{\textwidth}{!}{%
\begin{tabular}{lllrrrrrrrrr}
\toprule
 & & & \multicolumn{3}{c}{Base} & \multicolumn{3}{c}{AEGIS} & \multicolumn{3}{c}{\method{}} \\
\cmidrule(lr){4-6}\cmidrule(lr){7-9}\cmidrule(lr){10-12}
Backbone & Level & Task & SR $\uparrow$ & CC / $\mathrm{CC}_{\mathrm{policy}}$ $\downarrow$ & BRS$\uparrow$ & SR $\uparrow$ & CC / $\mathrm{CC}_{\mathrm{policy}}$ $\downarrow$ & BRS$\uparrow$ & SR $\uparrow$ & CC / $\mathrm{CC}_{\mathrm{policy}}$ $\downarrow$ & BRS$\uparrow$ \\
\midrule
\multicolumn{12}{l}{\textit{Safety / Static obstacles: pick the named object and place it in the bowl or plate}} \\
$\pi_{0.5}$ & 1 & T0 Apple & 90.0 & 9.12 / 0.12 & 0.368 & 82.0 & \textbf{8.20 / 0.00} & \textbf{0.407} & \textbf{90.3} & 9.21 / 0.21 & 0.261 \\
 & & T1 Lemon & \textbf{83.0} & 44.07 / 35.87 & 0.368 & 78.0 & \textbf{7.81 / 0.01} & \textbf{0.524} & 81.0 & 25.98 / 17.88 & 0.446 \\
 & & T2 Mango & 77.0 & 74.08 / 66.38 & 0.368 & 68.0 & \textbf{21.08} / 14.28 & 0.448 & \textbf{93.7} & 23.12 / \textbf{13.76} & \textbf{0.786} \\
 & & T3 Onion & 72.0 & 33.88 / 27.18 & 0.368 & 44.0 & \textbf{9.12 / 4.71} & 0.337 & \textbf{90.0} & 31.63 / 23.50 & \textbf{0.543} \\
 & & T4 Tomato & 85.0 & 45.31 / 37.01 & 0.368 & \textbf{94.0} & \textbf{17.39 / 7.99} & \textbf{0.735} & 93.3 & 34.28 / 25.04 & 0.571 \\
\midrule
 & 2 & T0 Apple & 8.0 & \textbf{67.51} / 65.89 & 0.368 & 7.3 & 90.73 / 89.26 & 0.307 & \textbf{34.2} & 68.75 / \textbf{61.90} & \textbf{0.437} \\
 & & T1 Lemon & \textbf{0.0} & 158.14 / 158.13 & 0.368 & \textbf{0.0} & 141.51 / 141.51 & 0.388 & \textbf{0.0} & \textbf{110.14 / 110.12} & \textbf{0.428} \\
 & & T2 Mango & 89.3 & 51.41 / 33.93 & 0.368 & 88.0 & \textbf{28.75 / 11.15} & 0.483 & \textbf{95.1} & 35.24 / 16.61 & \textbf{0.623} \\
 & & T3 Onion & 81.3 & 16.33 / 0.067 & 0.368 & 1.3 & \textbf{0.27 / 0.000} & 0.071 & \textbf{82.7} & 16.54 / 0.004 & \textbf{0.608} \\
 & & T4 Tomato & 74.0 & 67.87 / 53.07 & 0.368 & 28.7 & \textbf{26.93 / 21.20} & 0.208 & \textbf{84.9} & 55.44 / 38.55 & \textbf{0.520} \\
\midrule
$\pi_0$ & 1 & T0 Apple & 54.0 & \textbf{5.40 / 0.00} & -- & \textbf{82.0} & 8.20 / \textbf{0.00} & -- & 59.0 & 5.90 / \textbf{0.00} & -- \\
 & & T1 Lemon & 66.0 & \textbf{6.60 / 0.00} & -- & 80.0 & 8.00 / \textbf{0.00} & -- & \textbf{84.0} & 8.40 / \textbf{0.00} & -- \\
 & & T2 Mango & 92.0 & 11.34 / 2.14 & \textbf{0.368} & 82.0 & \textbf{9.40 / 1.20} & 0.245 & \textbf{93.0} & 12.39 / 3.09 & 0.314 \\
 & & T3 Onion & 28.0 & 18.34 / 16.74 & 0.368 & \textbf{30.0} & \textbf{3.00 / 0.00} & \textbf{0.615} & 24.0 & 12.02 / 10.32 & 0.433 \\
 & & T4 Tomato & 56.0 & 5.60 / \textbf{0.00} & -- & 44.0 & \textbf{4.40 / 0.00} & -- & \textbf{84.0} & 8.46 / 0.06 & -- \\
\midrule
 & 2 & T0 Apple & 0.0 & 66.31 / 66.31 & 0.368 & 0.0 & 54.63 / 54.62 & 0.402 & \textbf{0.3} & \textbf{50.29 / 50.22} & \textbf{0.416} \\
 & & T1 Lemon & 2.0 & 14.09 / 13.69 & 0.368 & \textbf{10.7} & 7.39 / 5.26 & \textbf{0.523} & 0.3 & \textbf{4.53 / 4.46} & 0.511 \\
 & & T2 Mango & \textbf{95.3} & 22.47 / 3.67 & 0.368 & 94.7 & \textbf{19.61 / 0.67} & \textbf{0.515} & 94.7 & 24.41 / 5.80 & 0.256 \\
 & & T3 Onion & \textbf{30.7} & 6.13 / \textbf{0.000} & -- & 6.7 & \textbf{1.34} / 0.007 & -- & 25.3 & 5.07 / \textbf{0.000} & -- \\
 & & T4 Tomato & 50.7 & 13.75 / 3.61 & 0.368 & 19.3 & \textbf{4.81 / 0.95} & \textbf{0.387} & \textbf{79.3} & 22.67 / 6.93 & 0.311 \\
\bottomrule
\end{tabular}
}
\end{table}

\subsection{Base-Relative Score: Definition and Limitations}
\label{app:score-b}

Equation~\ref{eq:score-b} combines two dimensionless quantities. The first is the method-to-base failure-rate ratio, and the second is the method-to-base policy-induced CC ratio. The two terms receive equal weight, so there is no fitted trade-off
parameter. The base score is $e^{-1}\approx0.368$, and an ideal policy with no
failure and no policy-induced contact scores 1. BRS is a base-relative
summary score. It is not a collision-avoidance guarantee or a standalone safety
metric.

Table~\ref{tab:score-b-bootstrap} reports 2,000 offset-level bootstrap samples.
The three methods share the same resampled offsets, and $\pi_{0.5}$ \method{}
first averages its three training orders within each offset on both levels. Confidence intervals
therefore quantify uncertainty in each task-specific score rather than treating
repeated sampler conditions as independent trials.

\begin{table}[htbp]
\centering
\caption{\textbf{Per-task BRS with 95\% bootstrap intervals.}
L2-T1 and L2-T4 were evaluated after the score was defined.}
\label{tab:score-b-bootstrap}
\footnotesize
\setlength{\tabcolsep}{4.0pt}
\begin{tabular}{lrrr}
\toprule
Task & Base & AEGIS $[95\%\ \mathrm{CI}]$ & \method{} $[95\%\ \mathrm{CI}]$ \\
\midrule
L1-T0 Apple  & 0.368 & \textbf{0.407} [0.109, 0.707] & 0.261 [0.004, 0.535] \\
L1-T1 Lemon  & 0.368 & \textbf{0.524} [0.299, 0.723] & 0.446 [0.250, 0.619] \\
L1-T2 Mango  & 0.368 & 0.448 [0.268, 0.613] & \textbf{0.786} [0.657, 0.888] \\
L1-T3 Onion  & 0.368 & 0.337 [0.199, 0.459] & \textbf{0.543} [0.427, 0.649] \\
L1-T4 Tomato & 0.368 & \textbf{0.735} [0.493, 0.926] & 0.571 [0.341, 0.754] \\
\midrule
L2-T0 Apple  & 0.368 & 0.307 [0.243, 0.363] & \textbf{0.437} [0.374, 0.492] \\
L2-T1 Lemon  & 0.368 & 0.388 [0.366, 0.408] & \textbf{0.428} [0.401, 0.452] \\
L2-T2 Mango  & 0.368 & 0.483 [0.173, 0.731] & \textbf{0.623} [0.359, 0.789] \\
L2-T3 Onion  & 0.368 & 0.071 [0.019, 0.147] & \textbf{0.608} [0.494, 0.711] \\
L2-T4 Tomato & 0.368 & 0.208 [0.113, 0.302] & \textbf{0.520} [0.416, 0.618] \\
\bottomrule
\end{tabular}
\end{table}

The score was selected after inspecting L1-T0--T4 and L2-T0, T2, and T3. L2-T1 and L2-T4 were evaluated afterward and provide checks on tasks whose outcomes were not used to select the score definition.
Using the three-order mean on both levels, \method{} attains higher BRS than
AEGIS on seven of ten $\pi_{0.5}$ tasks, including both held-out tasks.
AEGIS scores higher on L1-T0, L1-T1, and L1-T4.

Small base denominators make $\mathrm{BRS}$ unstable. L1-T0 has a base failure rate of
$10\%$ and base policy-induced CC of 0.12, which yields a wide interval for \method{}.
L2-T3 has base policy-induced CC of only 0.067, and only 1,273 of 2,000 bootstrap resamples have a nonzero cost denominator; BRS is undefined in the remaining resamples. On L2-T1, every method has zero SR,
so $\mathrm{BRS}$ is determined entirely by CC. BRS should therefore be interpreted alongside SR, official CC, and policy-induced CC, with its denominator sensitivity made explicit.

The $\pi_0$ comparison was evaluated after the definition of BRS was fixed,
so it also tests the score without backbone-specific retuning. The base
policy-induced CC is zero on L1-T0, L1-T1, L1-T4, and L2-T3. We retain dashes
for these undefined scores rather than substituting CC or adding a denominator
offset. Table~\ref{tab:pi0-aegis-score} reports the AEGIS intervals from the
completed cross-backbone comparison. These values should be read alongside
the task success rates, not as evidence of universal method dominance.

\begin{table}[htbp]
\centering
\caption{AEGIS BRS on $\pi_0$ tasks with defined normalization.}
\label{tab:pi0-aegis-score}
\small
\begin{tabular}{lrr}
\toprule
Task & Base $\mathrm{BRS}$ & AEGIS $\mathrm{BRS}$ [95\% CI] \\
\midrule
L1-T2 Mango & 0.368 & 0.245 [0.009, 0.607] \\
L1-T3 Onion & 0.368 & 0.615 [0.564, 0.663] \\
L2-T0 Apple & 0.368 & 0.402 [0.350, 0.446] \\
L2-T1 Lemon & 0.368 & 0.523 [0.438, 0.609] \\
L2-T2 Mango & 0.368 & 0.515 [0.209, 0.766] \\
L2-T4 Tomato & 0.368 & 0.387 [0.271, 0.502] \\
\bottomrule
\end{tabular}
\end{table}

Across ten tasks, BRS computed from unweighted mean SR and policy-induced CC is 0.498 for $\pi_{0.5}$ \method{}, with a bootstrap interval from 0.469 to 0.529. AEGIS scores 0.350, with an interval from 0.319 to 0.382. On $\pi_0$, the corresponding scores are 0.443 and 0.441, with intervals from 0.402 to 0.476 and from 0.401 to 0.475. The bootstrap probability that \method{} scores higher than AEGIS is 0.55.
The intervals overlap and do not support a BRS advantage on $\pi_0$.
Aggregate BRS uses suite-level mean denominators. Per-task BRS retains dashes
where the task-specific base denominator is zero. Neither comparison
establishes that \method{} is safer than AEGIS.

\subsection{Training and Evaluation Levels}
\label{app:benchmark-levels}

All arms inherit the Arena-published VLA checkpoints finetuned on L0
demonstrations. Level~0 is therefore the source difficulty represented in the
released checkpoints rather than a held-out test of feedback-driven improvement.
We use Level~1 to measure adaptation beyond that source and Level~2 to test
transfer to the hardest benchmark difficulty. \method{} then performs a distinct
self-evolution stage. Two observe-only rounds are collected on 47 L1
static-obstacle T2 cells and merged
into the main round-2 bank. No L2 rollout enters this bank, so every L2 result is
zero-shot with respect to the main policy-update data. A separate state-holdout
protocol trains on L1-T2 offsets 0--31 and evaluates 32--49. Its existing adapter
and base are evaluated on the same GPU host. On the 18 unseen initial states,
the host-matched comparison gives $93.3\%$ SR for \method{} and $71.1\%$ for base.
The policy-induced CC difference is not supported by the paired test.

\subsection{Backbone Scope}
\label{app:backbone-scope}

We audited the complete VLA-Arena leaderboard before selecting the reported
backbones. The audit covered 29 listed models, of which 10 released
Arena-finetuned checkpoints. A candidate had to satisfy three conditions. Each candidate required an available Arena checkpoint, a continuous flow-matching loss compatible with first-action supervision and the guard, and usable baseline task behavior with measurable headroom.
Table~\ref{tab:backbone-audit} records the resulting scope. Rows that group model
variants account for all 10 released checkpoints.

\begin{table}[htbp]
\centering
\caption{\textbf{Audit of Arena-finetuned backbone candidates.} Inapplicable
models are architectural exclusions rather than negative method results.}
\label{tab:backbone-audit}
\scriptsize
\setlength{\tabcolsep}{3.2pt}
\begin{tabular}{lclp{0.46\linewidth}}
\toprule
Model & Arena L1 SR & Action interface & Scope decision \\
\midrule
$\pi_{0.5}$-FFT & 0.64 & Flow matching & Main backbone with complete matched evaluation \\
$\pi_0$ / $\pi_0$-FFT & 0.74 / 0.76 & Flow matching & Second backbone with complete matched evaluation \\
$\pi_0$-FAST / -FFT & 0.40 / 0.60 & FAST tokens & Inapplicable because the chunk-level continuous action loss is absent \\
OpenVLA & 0.60 & Discrete autoregressive & Inapplicable to the current loss and guard \\
OpenVLA-OFT & 0.20 & Discrete autoregressive & Inapplicable to the current loss and guard \\
UniVLA & 0.42 & Autoregressive latent action & Inapplicable to the current loss and guard \\
SmolVLA & 0.00 & Flow matching & Excluded because the Arena baseline has no viable L1 task behavior \\
LangForce & 0.72 & Diffusion & Outside the shared openpi training and serving stack \\
\bottomrule
\end{tabular}
\end{table}

The architectural boundary is clearest for $\pi_0$-FAST
\citep{pertsch2025fastefficientactiontokenization}. Its Arena checkpoint loads
successfully and its baseline is not saturated, with 91 of 150 audited cells
remaining improvable. However, its parameter tree has 32 leaves instead of the
50 leaves in $\pi_0$ and contains none of the continuous action-expert
parameters used by our update. Its loss reduces the token axis to one scalar per batch element. The present supervision requires a loss indexed by action-chunk step. The attempted update therefore stops at the first loss-shape check before
an adapter or evaluation result is produced. OpenVLA and UniVLA are excluded for
the same objective-level incompatibility
\citep{kim2024openvlaopensourcevisionlanguageactionmodel,
bu2025univlalearningacttaskcentric}. SmolVLA exposes a flow-matching interface,
but its released Arena checkpoint has zero L1 success and uses a different
in-process serving path
\citep{shukor2025smolvlavisionlanguageactionmodelaffordable}.

Transferring the recipe from $\pi_{0.5}$ to $\pi_0$ also required a stronger
quiet-record regularizer. The $\pi_0$ update passed the fixed guard at
a quiet-anchor weight of 0.5, whereas weights from 0 to 0.3 did not. Extending supervision
from the first action to 10 or 50 chunk steps reduced the guard loss ratio but
did not improve five-task success, indicating that guard passage alone was not
sufficient. The reported method comparison is therefore restricted to the two backbones with a complete matched update-and-evaluation chain.

The completed $\pi_0$ comparison adds AEGIS on all five Level 1 tasks. Each task
uses 50 initial states. The comparison also evaluates base, AEGIS, and the
guarded adapter on all five Level 2 tasks. Each Level 2 task uses 50 states with
three sampler conditions. The 3,500
added evaluation cells pass the host and architecture checks, and all 1,000
AEGIS cells report successful perception. The restored adapter checkpoint
matches the audited source checkpoint in every cell. Level 1 base and adapter
references share qa-l40s-004 with the added AEGIS arm, while Level 2 comparisons
use their matched task hosts. This audit establishes comparison provenance, not
an attribution of gains to shield corrections. The $\pi_0$ Level 2 apple and
lemon rows satisfy the registered floor criterion and are reported without
method conclusions.

\subsection{Task Coverage and Names}
\label{app:task-coverage}

Table~\ref{tab:coverage-ledger} distinguishes matched method comparisons, held-out evaluations, and exploratory base-only probes. Task indices are interpreted within their suite and difficulty level.

\begin{table}[htbp]
\centering
\caption{\textbf{Complete ledger of reported and exploratory task coverage.}
Base-only checks are not method comparisons.}
\label{tab:coverage-ledger}
\small
\setlength{\tabcolsep}{3.5pt}
\begin{tabular}{llll}
\toprule
Evidence & Suite / level & Tasks & Protocol \\
\midrule
Main static & static obstacles / L1 & T0--T4 & base, AEGIS, \method{}, 50 states \\
State holdout & static obstacles / L1 & T2 & train 32 states, test 18 unseen states \\
Zero-shot static & static obstacles / L2 & T0--T4 & base, AEGIS, 3 \method{} orders \\
Static rounds & static obstacles / L1 & T3 & base and R1--R5, 50 states \\
Dynamic stress test & dynamic obstacles / L1 & T3, T4 & base and learned policies, 18 states \\
Dynamic rounds & dynamic obstacles / L2 & T0 & base and R1--R3, 50 states \\
Backbone transfer & static obstacles / L1, L2 & T0--T4 & $\pi_0$ base / AEGIS / \method{} \\
Exploratory checks & 3 other Safety suites / L2 & all T0--T4 & base only, 20 states \\
\bottomrule
\end{tabular}
\end{table}

Static-obstacle task indices T0--T4 manipulate apple, lemon, mango, onion, and
tomato, respectively, placing the named object in a bowl or plate while avoiding
a protected external object. Dynamic-obstacle T0 picks and places an apple.
T1--T4 push lemon, onion, peach, and tomato, respectively, while an obstacle can
move through the workspace. Thus ``hard apple,'' ``mango,'' and ``onion'' in the
main text are task names within a fixed suite and level, not new metrics.

The original Level 2 evaluation used T0, T2, and T3 to diagnose AEGIS obstacle
selection under heterogeneous, earlier-tested scene configurations. It was not
designed as a coverage sample. Before reading the remaining outcomes, we added
T1 and T4 under the same 50-offset, three-condition protocol and registered the
decision rule. T4 reproduces the zero-shot transfer gain in two of three
training orders with no significant regression in any order. T1 satisfies the
registered floor criterion because base and all three updated policies obtain
zero SR, so we report it for coverage without drawing a method conclusion.

\subsection{Generalization Across States, Tasks, and Levels}
\label{app:generalization-radii}

Table~\ref{tab:generalization-rings} separates unseen initial states, tasks, and difficulty levels. The state-holdout adapter trains on L1-T2 initial-state offsets 0--31 and is evaluated on offsets 32--49. The main cross-task and Level~2 evaluations use the L1-T2 collection bank described in Appendix~\ref{app:benchmark-levels}. No Level~2 rollout enters that bank.

\begin{table}[htbp]
\centering
\caption{\textbf{Three widening generalization radii.} Only L1-T2 rollouts enter
the self-evolution bank. Split details are given above.}
\label{tab:generalization-rings}
\small
\setlength{\tabcolsep}{5pt}
\resizebox{\textwidth}{!}{%
\begin{tabular}{llll}
\toprule
Radius & Training exposure & Evaluation & Result \\
\midrule
Unseen states & L1-T2 training states & held-out L1-T2 states & SR $71.1\%\!\rightarrow\!93.3\%$ \\
Unseen tasks & L1-T2 only & other Level 1 tasks on $\pi_{0.5}$ & mean SR and paired test in Table~\ref{tab:breadth-paired} \\
Unseen level & Level 1 only & Level 2 T0--T4 & apple/tomato gains, mango/onion preserved, lemon floor \\
\bottomrule
\end{tabular}
}
\end{table}

\subsection{Scope Boundary}
\label{app:scope-boundary}

Our complete three-arm method evaluation uses the static-obstacle suite.
VLA-Arena has eleven task categories. They comprise five Safety, two Distractor,
three Extrapolation, and one Long-Horizon category, in addition to five original
LIBERO suites. All five Safety suites define a cost predicate, but their
constraints differ substantially. Static Distractor, all three Extrapolation
categories, Long Horizon, and the five LIBERO suites contain no
benchmark cost predicate. Dynamic Distractor has contact costs but was not evaluated.
Its moving hazards share the snapshot-geometry issue examined in the dynamic
stress tests. Table~\ref{tab:scope-audit} preserves the complete category audit.

\begin{table}[htbp]
\centering
\caption{\textbf{Evaluation-scope audit from BDDL predicates and measured probes.}
The count column gives the number of tasks with a cost predicate followed by
the number of tasks inspected. A missing cost
predicate provides no benchmark cost axis. It is not a negative method result.}
\label{tab:scope-audit}
\scriptsize
\setlength{\tabcolsep}{3pt}
\begin{tabular}{p{0.10\linewidth}p{0.17\linewidth}p{0.18\linewidth}p{0.06\linewidth}p{0.23\linewidth}p{0.17\linewidth}}
\toprule
Category & Suite & Cost semantics & Count & Measured evidence & Evaluation status \\
\midrule
Safety & Static obstacles & Fall and contact with protected external bodies & 10/15 & Complete matched three-arm results & Main method comparison \\
Safety & Dynamic obstacles & Contact with moving bodies and fall & 15/15 & Base L1-T0 SR 95\%, dynamic transfer fails & Negative stress test \\
Safety & Cautious grasp & Part-level gripper distance to target itself & 15/15 & Base SR 0/100. Oracle hazard set empty & Floor and unrepresented constraint \\
Safety & State preservation & Containment predicates & 10/15 & Base SR 66/100. Zero action rewrites in 196 AEGIS chunks & Incompatible avoidance semantics \\
Safety & Hazard avoidance & Surface-distance dwell cost near stove/candle and fall & 15/15 & Initial objects in cost region 42--50/50. VLM misidentifies both preliminary validity cells & Teacher pilot and AEGIS validity gate fail \\
Distractor & Static distractors & No cost predicate & 0/15 & BDDL audit & No benchmark cost axis \\
Distractor & Dynamic distractors & Contact with moving toys & 10/15 & Not evaluated. Motion lasts 50--75 steps, 0.5--0.65 m travel & Unevaluated moving-hazard suite \\
Extrapolation & Preposition combinations / task workflows / unseen objects & No cost predicate & 0/45 & BDDL audit, 15 tasks per category & No benchmark cost axis \\
Long Horizon & Long horizon & No cost predicate & 0/20 & BDDL audit & No benchmark cost axis \\
LIBERO & Spatial / object / goal / 10 / 90 & No cost predicate & 0/130 & BDDL audit & No benchmark cost axis \\
\bottomrule
\end{tabular}
\end{table}

An external-body contact-avoidance teacher represents the static/dynamic
obstacle constraints. Cautious Grasp instead constrains the manipulated object's
parts, and State Preservation constrains containment. The latter's extracted
``hazard'' is the water itself, so repelling it does not preserve containment.
Hazard Avoidance counts dwell time in a region occupied by the task's starting
object and often its destination. Among the obstacle suites, the tested dynamic
baseline has little headroom and the method fails to transfer. These observations
motivate the static-obstacle comparison. They do not establish that the method
works in every static scene or fails on every untested suite. CC is compared
only within the same suite and cost semantics.

\subsubsection{Hazard-Avoidance Validity and Headroom}
\label{app:hazard-validity}

Hazard Avoidance charges each step for an object or gripper lying within a
fixed surface distance of a stove or candle. Fall is evaluated at episode end.
The object already lies in the cost region in 42--50 of 50 initial states per
task, and several destination containers lie there as well. A repelling barrier
therefore conflicts with grasping or placing the object. The lift-gated
withdrawal-teacher pilot leaves L2-T0 SR unchanged at 48.0\%. It reduces
L2-T4 SR from 48.0\% to 8.0\%. The paired test gives a $p$-value of $0.002$. Its lower L2-T4 cost reflects
failed completion, not safer successful manipulation. It fails the preregistered
two-task gate, so no \method{} update is trained for this suite.

The completed base evaluation covers both backbones, two levels, and all five
tasks per level. This design covers 20 base-policy task arms and 1,000 initial-state offsets. These base-only results do not constitute a paired method comparison. Ten arms have base
SR of at most $5\%$. AEGIS is checked only in two preliminary validity cells, $\pi_0$
L1-T0 for the candle task and L1-T1 for the stove task. Both have successful
perception status and nonempty cropped point clouds. The point clouds contain
626 and 4,243 points, respectively. The shield activates
constraints on 9 and 31 steps, with no infeasible QP, but GLM-4.5V identifies
the obstacle as ``black wine bottle'' in both cells. This fails the preregistered
semantic gate requiring stove or candle identification. AEGIS is therefore
not run in a full comparison on this suite. This is a grounding failure rather
than a failed perception crop. Table~\ref{tab:hazard-base} reports the base
results without implying a method comparison. Hazard-Avoidance cost values
are not pooled with static-obstacle costs.

\begin{table}[htbp]
\centering
\caption{\textbf{Hazard-Avoidance base-policy headroom.} The evaluation includes
all 20 base-policy task arms and 50 offsets per arm. CC is the archived official cost; the final column reports the archived policy-induced cost. These columns use different scales and cannot be interpreted as the additive decomposition in Equation~\ref{eq:metrics-policy}. Cost-unit reconciliation is required before comparing them.}
\label{tab:hazard-base}
\footnotesize
\begin{tabular}{llrrr|rrr}
\toprule
Level & Task & \multicolumn{3}{c|}{$\pi_{0.5}$ base} & \multicolumn{3}{c}{$\pi_0$ base} \\
 & & SR & CC & Logged policy cost & SR & CC & Logged policy cost \\
\midrule
L1 & T0 & 9.0 & 16.75 & 334.9 & 4.0 & 15.99 & 319.7 \\
 & T1 & 0.0 & 21.55 & 430.6 & 2.0 & 16.68 & 333.2 \\
 & T2 & 62.0 & 8.09 & 161.8 & 2.0 & 11.12 & 222.3 \\
 & T3 & 34.0 & 15.79 & 315.6 & 12.0 & 20.36 & 406.8 \\
 & T4 & 0.0 & 20.67 & 413.5 & 0.0 & 22.40 & 448.0 \\
\midrule
L2 & T0 & 38.7 & 14.20 & 284.0 & 0.7 & 17.58 & 350.7 \\
 & T1 & 14.0 & 19.91 & 397.8 & 0.0 & 18.39 & 366.9 \\
 & T2 & 21.3 & 14.41 & 288.0 & 0.7 & 19.83 & 396.1 \\
 & T3 & 48.7 & 16.51 & 330.3 & 0.7 & 22.67 & 453.3 \\
 & T4 & 40.0 & 11.61 & 232.2 & 76.0 & 12.70 & 254.1 \\
\bottomrule
\end{tabular}
\end{table}

\subsection{Dynamic-Obstacle Stress Tests}
\label{app:dynamic-tests}

A base-only L1-T0 probe reaches 95\% SR, with zero policy-induced CC on
19 of 20 cells, providing evidence of limited baseline headroom.
We test whether the static L1-T2 update transfers to dynamic
obstacles. On 18 matched L1-T3 states, base and the main adapter reach $55.6\%$
and $72.2\%$ SR, but the paired result is inconclusive. On L1-T4 both reach
$5.6\%$ SR. Pooled across the two tasks, the main adapter records 6 improvements,
3 regressions, and 27 ties. The paired test gives a $p$-value of $0.5078$. These results do not establish transfer
from the static teacher to dynamic tasks.

A separate diagnostic adapter trained on dynamic L2-T0 data reaches $72.2\%$ on
L1-T3 and $16.7\%$ on L1-T4. Neither task-level comparison is supported by the
paired tests, so these rows are not included in the paper's generalization pool.
The controlled L2-T0 round study is also null against base. Matching the first
round to the later 700-step budget leaves it at $86.0\%$ SR, significantly above
round two at $70.0\%$. The comparison contains 10 improvements and 2
regressions, and the paired test gives a $p$-value of $0.0386$. The decline persists after matching the update budget, so a budget difference alone does not explain it.

At the short horizons consumed by the dynamic barrier, first-order extrapolation
error is only $10$--$29\%$ of the oracle hazard radius even with simulator-truth
velocity. This diagnostic indicates limited prediction headroom at the measured horizons when simulator-truth velocity is available. It does not test learned perception or longer-horizon planning, which the current barrier interface does not consume.


\section{Collection and Failure-Bank Analysis}
\label{app:collection}

\subsection{Collection-Mode Outcome Counts}
\label{app:collection-outcomes}

\begin{figure}[htbp]
\centering
\includegraphics[width=\textwidth]{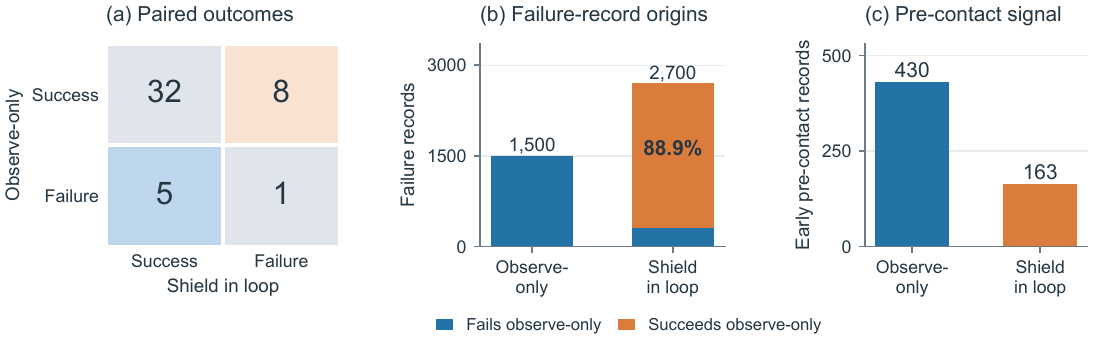}
\caption{\textbf{Observe-only versus shield-in-loop collection on Level~1 mango.} Panel a shows paired episode outcomes, with observe-only outcomes in rows and shield-in-loop outcomes in columns. Panel b groups failure records by paired episode outcome. Panel c counts stored early pre-contact records before outcome-aware admission.}
\label{fig:collection-analysis}
\end{figure}

Figure~\ref{fig:collection-analysis} compares observe-only and shield-in-loop
collection on the same Level 1 mango initial states. Across the 46 paired cells,
32 succeed under both collection modes, five failures are rescued by steering,
eight successes become failures under steering, and one cell fails under both.
The comparison shows that executing the shield changes the trajectory outcomes
from which learning records are collected.

\subsection{Failure-Record Provenance}
\label{app:collection-provenance}

Observe-only and shield-in-loop collection produce 5,801 and 7,728
training-fold records, respectively. The in-loop bank is therefore larger, but
its additional records do not necessarily correspond to failures of the
base policy.

Of the 2,700 failure records in the shield-in-loop bank, 2,400 come from the
eight initial states that succeed without steering but fail when
the shield is executed. These are failures of the shielded system rather than
failures encountered under the policy's original state distribution.
These records constitute $88.9\%$ of the 2,700 failure records, not of the full 7,728-record training fold. This record-weighted percentage
is neither the fraction of episodes that fail nor an adapter-level effect size.

\subsection{Pre-Contact Correction Signal}
\label{app:collection-precontact}

Steering also changes the stage at which useful correction records are
observed. Early pre-contact records decrease from 430 under observe-only
collection to 163 with the shield in the loop, a reduction of approximately
$62\%$, as shown in Figure~\ref{fig:collection-analysis}c.

The lead-time counts describe the stored bank before outcome-aware record
admission and therefore should not be interpreted as the number of admitted
corrective targets. The provenance and pre-contact analyses explain why observe-only collection retains evidence from trajectories controlled by the policy being updated.

\subsection{Adapter-Level Collection Ablation}
\label{app:collection-adapter}

The two first-round banks are collected from the same policy checkpoint and
trained with the same 800-step recipe with zero quiet-anchor weight. The observe-only bank
contains 3,657 training records and passes the held-out guard. Its adapter raises
Level 2 apple SR from $8.0\%$ to $34.0\%$. The in-loop bank contains 5,099 records
but no candidate checkpoint satisfies both guard conditions, so no deployable
adapter is retained. This comparison uses one training order, and its held-out batches differ in size. The observe-only batch contains 300 records, whereas the in-loop batch
contains 112 records.

\begin{table}[htbp]
\centering
\caption{Collection-mode training outcome under the main update recipe.}
\label{tab:collection-adapter}
\small
\begin{tabular}{lrrlc}
\toprule
Collection & Train records & Guard records & Guard decision & Apple SR \\
\midrule
Observe-only & 3,657 & 300 & pass & 34.0 \\
Shield in loop & 5,099 & 112 & reject & unavailable \\
\bottomrule
\end{tabular}
\end{table}

\subsection{Attribution Controls}
\label{app:attribution}

Only 709 of the 6,535 main-bank training targets carry a nonzero shield residual.
SFT0 replaces those residuals with zero, SHAM randomizes their direction while
preserving magnitude, and SFTPOS trains only on successful episodes. All other
aligned fields remain unchanged for SFT0 and SHAM. Table~\ref{tab:attribution-l2}
shows that \method{} outperforms SHAM and SFTPOS in every training order. It
outperforms SFT0 in two orders, while the third is inconclusive. Averaging the three training orders within each offset gives a $p$-value of $0.0011$ against SFT0, a $p$-value of $0.0029$ against
SHAM, and a $p$-value below $0.0001$ against SFTPOS.

\begin{table}[htbp]
\centering
\caption{Level 2 apple SR attribution controls over three training orders.}
\label{tab:attribution-l2}
\small
\begin{tabular}{lrrrr}
\toprule
Update target & Seed 1 & Seed 2 & Order 3 & Mean \\
\midrule
Base & 8.0 & 8.0 & 8.0 & 8.0 \\
SFTPOS & 4.0 & 10.7 & 10.7 & 8.4 \\
SFT0 & 18.0 & 14.7 & 25.3 & 19.3 \\
SHAM & 26.0 & 16.0 & 18.0 & 20.0 \\
\method{} & \textbf{36.0} & \textbf{31.3} & \textbf{35.3} & \textbf{34.2} \\
\bottomrule
\end{tabular}
\end{table}

\begin{table}[htbp]
\centering
\caption{\textbf{Level 1 attribution controls for one training order.}
Each cell reports SR followed by policy-induced CC over 50 offsets under two sampler
conditions.}
\label{tab:attribution-l1}
\footnotesize
\setlength{\tabcolsep}{4.2pt}
\begin{tabular}{lrrrrr}
\toprule
Task & Base & \method{} & SFT0 & SHAM & SFTPOS \\
\midrule
L1-T1 Lemon & 83.0 / 35.87 & 83.0 / 12.80 & 84.0 / 5.66 & 94.0 / 7.75 & 81.0 / 29.01 \\
L1-T4 Tomato & 85.0 / 37.01 & 90.0 / 31.24 & 86.0 / 33.51 & 82.0 / 40.61 & 81.0 / 51.27 \\
\bottomrule
\end{tabular}
\end{table}

The null updates themselves improve over base on Level 2 apple. SFT0 accounts
for 11.3 of the full 26.2-point gain and SHAM for 12.0 points. On Level 1 T1,
SHAM reaches $94.0\%$ SR versus $83.0\%$ for \method{} and achieves a comparable
cost reduction. On T4, the aligned comparisons are inconclusive. The evidence
therefore supports partial attribution on the hard apple task, not a universal
claim that shield residuals alone cause the gain.

An earlier attribution sweep used a different 2,830-record recipe and stopped
after 25--50 of 400 planned steps. It had no same-host base arm and predates the
cost-decomposition fix, so only its SR is interpretable. The ordering of the
method and SHAM reversed across seeds, and cross-task transfer was inconclusive.
We report this audit to distinguish the current matched controls from exploratory
development results.

\subsection{Privileged Teacher Executed In Loop}
\label{app:teacher-loop}

To separate projection effects from perception errors, we execute the same
privileged-geometry CBF used to label records. Table~\ref{tab:teacher-loop}
reports 50 offsets under three sampler conditions per task. The shield-in-loop
arm collapses onion to zero success, reduces apple to zero, and sharply degrades
mango while increasing its cost. Its outcome differs from the matched base in
$70\%$, $87\%$, and $82\%$ of apple, mango, and onion cells, respectively,
showing that the two arms produced different outcomes.

\begin{table}[htbp]
\centering
\caption{Executing the privileged CBF teacher as a shield. Cost cells report
CC followed by policy-induced CC.}
\label{tab:teacher-loop}
\small
\begin{tabular}{llrrr}
\toprule
Task & Method & SR & CC & $\mathrm{CC}_{\mathrm{policy}}$ \\
\midrule
Apple & Base & 8.0 & 67.51 & 65.89 \\
 & AEGIS & 7.3 & 90.73 & 89.26 \\
 & Teacher in loop & 0.0 & 67.02 & 67.01 \\
\midrule
Mango & Base & 89.3 & 51.41 & 33.93 \\
 & AEGIS & 88.0 & 28.75 & 11.15 \\
 & Teacher in loop & 50.7 & 107.63 & 97.50 \\
\midrule
Onion & Base & 81.3 & 16.33 & 0.07 \\
 & AEGIS & 1.3 & 0.27 & 0.00 \\
 & Teacher in loop & 0.0 & 0.00 & 0.00 \\
\bottomrule
\end{tabular}
\end{table}

The logged correction and trigger counters remain zero on this execution path, so they cannot verify whether individual projections were executed. We instead use the pre-registered matched-outcome
divergence check. The privileged teacher is not better than AEGIS on any of the
three tasks. Collapse also occurs with privileged geometry, so these runs do not require perception errors to explain the loss of task completion. The inactive counters do not directly verify individual projections.


\section{Deployment Boundary}
\label{app:deployment}

\subsection{Deployment Requirements}
\label{app:deployment-requirements}

\begin{table}[htbp]
\centering
\caption{Deployment requirements. Privileged object geometry is an offline
teacher input for \method{} and is absent at evaluation.}
\label{tab:resources}
\small
\begin{tabular}{lccccc}
\toprule
Method & Updated VLA & Shield & Object geometry & QP/step & s/step \\
\midrule
Base policy & no & no & no & no & 0.362 \\
AEGIS & no & yes & yes & yes & 0.449 \\
\method{} & LoRA & no & no & no & 0.346 \\
\bottomrule
\end{tabular}
\end{table}

The distinction in Table~\ref{tab:resources} is between information available
during learning and components required after the update. \method{} uses
privileged object geometry only to construct counterfactual teacher targets
during collection. Once the guarded LoRA adapter is accepted, deployment
requires only the updated VLA policy. It does not use a runtime shield,
privileged geometry, a quadratic program, a retrieval system, a memory lookup,
or a test-time update.
The latency audit uses the first ten Level 2 apple offsets, one concurrent run per
arm, and reports the median total wall clock divided by executed steps. It
includes policy-server startup and checkpoint restoration, so it supports only a
matched relative comparison. The observed median for folded \method{} is close to base. AEGIS adds approximately $24\%$ and serves a 202~GB vision--language model in addition to its per-step QP. This wall-clock audit does not establish isolated inference latency or a speedup from adaptation.


\section{Additional Training and Self-Evolution Diagnostics}
\label{app:diagnostics}

\subsection{Implementation and Training Details}
\label{app:implementation}

Table~\ref{tab:hyperparameters} lists the shared training configuration. LoRA is
inserted into the PaliGemma backbone and the flow-matching action expert, while
all non-LoRA parameters remain frozen. The main adapter uses training-order seed
0, and the three Level 2 replicates use seeds 1--3.

\begin{table}[htbp]
\centering
\caption{Policy-update hyperparameters.}
\label{tab:hyperparameters}
\small
\begin{tabular}{ll}
\toprule
Component & Setting \\
\midrule
PaliGemma 2B LoRA & rank 16, $\alpha=16$ \\
300M action-expert LoRA & rank 32, $\alpha=32$ \\
Trainable parameters & LoRA $A/B$ factors only \\
Optimizer & AdamW, gradient clipping 1.0 \\
Learning-rate schedule & cosine, 20-step warmup, peak $3\!\times\!10^{-5}$ \\
 & 400-step decay, final $3\!\times\!10^{-6}$ \\
Batch size & 32 \\
Action horizon & 10 for $\pi_{0.5}$ and 50 for $\pi_0$ \\
Replanning / episode limit & every step / 300 control steps \\
Main update & 800 steps, $\lambda_q=0$ \\
Round-curve family & 800 steps, $\lambda_q=0.2$ \\
Cross-backbone $\pi_0$ & 800 steps, $\lambda_q=0.5$ \\
\bottomrule
\end{tabular}
\end{table}

\subsubsection{Barrier Teacher}
\label{app:barrier-teacher}
The teacher models protected object $j$ as an ellipsoid with center $c_j$ and positive-definite shape matrix $P_j$. The center and end-effector position $x$ are three-dimensional, and $P_j$ is a $3\times3$ matrix. The barrier is
\begin{equation}
 h_j(x)=(x-c_j)^\top P_j^{-1}(x-c_j)-1.
\end{equation}
For nominal translation $u\in\mathbb{R}^3$ and candidate translation $v\in\mathbb{R}^3$, the teacher solves
\begin{equation}
 \tilde u=\arg\min_v \tfrac12\lVert v-u\rVert_2^2
 \quad\text{s.t.}\quad \dot h_j(x,v)+3h_j(x)\geq0\quad\forall j.
\end{equation}
The end-effector radius is 0.03, protected-object inflation is 0.04, and the
additional margin is 0.02. The safety-region detector uses near and release
thresholds 0.205 and 0.307, with minimum approach speed 0.0027 m per step.
The default record-labeling implementation uses privileged geometry. AEGIS
instead calls GLM-4.5V once per episode to identify obstacles and then performs
the QP projection at every control step.

\subsubsection{Outcome-Aware Record Weight}
\label{app:teacher-weight}
For a triggered, feasible record with finite action and nonzero correction, the
outcome-aware score is
\begin{align}
q_i={}&0.40s_i+0.30r_i+0.20R_i+0.10(1-\rho_i)\\
&-0.30C_i-0.20F_i,
\end{align}
clipped to the range from zero to one. Here $s_i$ is normalized one-step forward safety, $r_i$ is
progress preservation, $R_i$ indicates observed recovery, $\rho_i$ is the
subsequent repeated-trigger rate, $C_i$ indicates four-step cost, and $F_i$
indicates a four-step barrier-floor violation. The gate additionally requires a
nonnegative one-step barrier or observed recovery. Records that fail the gate or score below 0.25 receive zero weight. Scores of at least 0.25 but below 0.45 receive weight 0.25. Scores of at least 0.45 but below 0.70 receive weight 0.60, and scores of at least 0.70 receive weight 1.00.

\subsubsection{Record Admission}
\label{app:record-admission}
The static L1-T2 risk threshold is 0.1087. Records are assigned to five
lead-time bins. Early pre-contact records occur more than 30 steps before
crossing. Mid pre-contact records occur 15--30 steps before crossing, and
emergency records occur 0--15 steps before crossing. The remaining bins are
post-crossing and no-risk. Emergency and post-crossing corrections are discarded.
On failed trajectories, only early pre-contact correction records are retained.
Successful actions without a triggered correction provide quiet anchors. A pre-training audit rejects
failure targets identical to the nominal action. The builder runs in the early-
and mid-stage mode, while failed trajectories remain restricted to early
pre-contact records.

Admitted CBF-triggered records receive the outcome-aware weights specified in Appendix~\ref{app:teacher-weight}. Quiet anchors receive the per-record weight specified by the update recipe in Appendix~\ref{app:recipe}.

\subsection{Bank and Evaluation Ledger}
\label{app:bank-ledger}

\begin{table}[htbp]
\centering
\caption{Learning-record banks used by the reported experiments. CBF-triggered and quiet counts refer to records with and without a triggered correction in the filtered training fold. They are not counts of corrective and nominal targets.}
\label{tab:bank-ledger}
\footnotesize
\resizebox{\textwidth}{!}{%
\begin{tabular}{llllll}
\toprule
Bank & Collection suite & Train records & CBF-triggered / quiet & Validation fold & Used by \\
\midrule
Main static two-round & static L1-T2 & 6,535 & 2,863 / 3,672 & 600 & main tables, L2, R2 ablations \\
Pre-filter static pool & static L1-T2 & 12,343 & 4,485 / 7,858 & 600 & audit only \\
State holdout & static L1-T2 & 4,006 & 1,799 / 2,207 & 600 & unseen-state evaluation \\
Static R1--R5 & static L1-T2 & 3,657 / 6,535 / 15,707 / 19,051 / 22,732 & -- & -- & round curve \\
Dynamic diagnostic R1 & dynamic L2-T0 & 3,192 & 642 / 2,550 & 139 & dynamic L1 stress test \\
Dynamic diagnostic R2 & dynamic L2-T0 & 9,898 & 1,818 / 8,080 & 266 & dynamic round analysis \\
$\pi_0$ transfer & static L1-T2 & 2,875 & 747 / 2,128 & -- & Table~\ref{tab:main} \\
\bottomrule
\end{tabular}}
\end{table}

The main static two-round bank contains 3,657 records from the first collection
round and 2,878 from the second. These are data-collection rounds, while every
reported main policy is fitted afresh from the same original base. Runtime logs,
not directory arm names, determine the effective checkpoint and method used by
each result.

\subsection{Host Matching and Hardware Confounding}
\label{app:hardware}

A historical audit split 150 $\pi_0$ baseline cells across two hosts. A total
of 124 cells achieved $66.9\%$ SR on one host, and 26 achieved $96.2\%$ on the other.
These were different offset subsets, not paired replays. The 29.3-point gap
does not isolate a causal hardware effect. Every comparison in the paper is
pinned to one hostname, and a runtime assertion rejects a cell from an incompatible host.

\subsection{Update Recipes and Training Budgets}
\label{app:recipe}

The main $\pi_{0.5}$ update uses 800 optimization steps with zero quiet-anchor weight. The static round-curve family uses the same budget with quiet-anchor weight 0.2. The cross-backbone $\pi_0$ update uses 800 steps with quiet-anchor weight 0.5. Table~\ref{tab:hyperparameters} lists the shared configuration.

The main round-2 training fold contains 2,863 triggered and 3,672 quiet records,
or 6,535 total. The 12,343-record pool is the pre-filter set and is not used
directly for training. A quiet-anchor weight of 0.2 applies to each record. Multiplying the quiet-record count by this weight and dividing by the triggered-record count gives a nominal ratio of 0.257. The value 0.2 therefore does not specify the weight of the quiet class as a whole. Stored outcome-aware record weights
further determine the actual denominator in Equation~\ref{eq:objective}.

On the 9,898-record dynamic two-round bank, 700 update steps pass with a
held-out full-chunk flow-loss ratio of 1.0985, whereas 800 steps fail at 1.1016. The main
6,535-record static bank passes at 800 steps with ratio 1.0065. Increasing quiet
weight can restore the ratio while inflating triggered loss by up to $48\times$.
These checks motivate the guarded update rather than treating additional
optimization as uniformly beneficial.

\subsection{Bank-Construction Audit}
\label{app:bank-audit}

An earlier round-three construction retained only 3,364 new-round records,
whereas the rebuilt union contains 15,707 records. We exclude the
replacement-bank run from the accumulated curve.

The rebuilt adapter passes the held-out guard with a flow-loss ratio of 0.9993,
but this training-side observation does not isolate a task-performance benefit
of retaining history. In the review-time size control, the round-2-only bank is
rejected while a size-matched accumulated bank passes. Once the guard is
disabled for diagnosis, however, their Level 2 apple SR values are 38.7 and
41.3 and are not significantly different. The first-round, size-matched
accumulated, and full accumulated adapters likewise reach 34.0, 41.3, and 36.0
SR on that task without supported pairwise differences. The only supported
benefit is lower L1-T1 policy-induced CC for the size-matched accumulated bank than for
the first-round bank, 10.61 versus 24.00. We therefore treat accumulation as a
record-retention mechanism and do not claim that a larger bank improves
success. The controlled multi-round result uses only reconstructed accumulated
banks.

\subsection{Constraint-Activation Analysis}
\label{app:activation}

Observe-only activation measures how often a proposed action triggers the
constraint without changing the trajectory. On 50 dynamic offsets, activation
rises from $17.74\%$ at base to $19.47\%$ after round one
and the paired test gives a $p$-value of $0.0328$. Activation then falls to $16.04\%$ and
$14.36\%$ after rounds two and three. Round three is $19\%$ below base. The
paired directions include 40 decreases and 10 increases, and the paired test
gives a $p$-value below $10^{-4}$. Contact-step, success, and cost changes against base remain
null.

\begin{figure}[htbp]
\centering
\includegraphics[width=\textwidth]{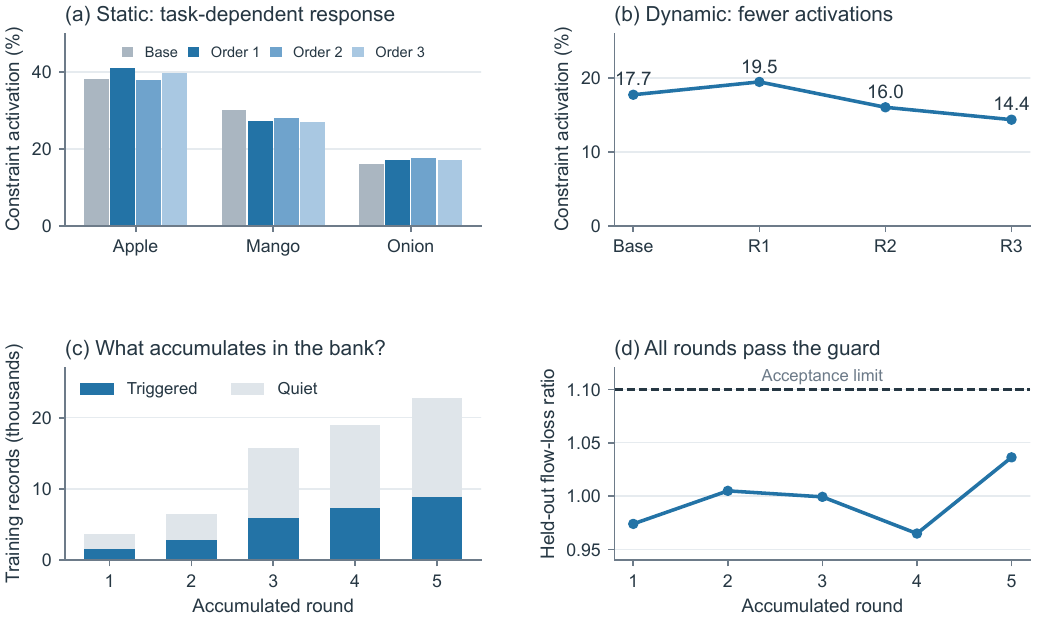}
\caption{\textbf{Training and self-evolution diagnostics.} Panels a and b show observe-only constraint activation for static tasks and dynamic rounds. Panel c separates accumulated training records by trigger provenance. Panel d shows held-out full-chunk flow-loss ratios relative to base. The dashed line marks the acceptance limit. Activation is a diagnostic, not a task-success or cost metric.}
\label{fig:diagnostic-panels}
\end{figure}

The direction is task-dependent. Mango activation decreases in all three
training orders. Each order has 36 lower and 14 higher offsets, and each paired
test gives a $p$-value of $0.0026$. Onion activation increases in two orders, with
a $p$-value of $0.0153$ and a $p$-value of $0.0066$. Apple shows mixed changes.
Figure~\ref{fig:diagnostic-panels}a--b shows all four tasks, including the
adverse round-one dynamic change. All comparisons are observe-only. None uses
the steering AEGIS arm's non-comparable counters.

\subsection{Held-Out Guard Diagnostics}
\label{app:guard-diagnostics}

The acceptance thresholds are fixed empirical optimization guardrails. They are
not theoretical CBF constants or statistically calibrated confidence bounds. The flow-loss ratio limit is 1.10, permitting at most a $10\%$ increase in full-chunk held-out loss relative to the original base policy. The first-action drift limit is 0.05, measured as the mean absolute difference per coordinate. Exceeding either value rejects the
adapter. Task SR and CC are not used for adapter acceptance, and the experiments
do not identify these thresholds as optimal.

The round-2 validation fold has 474 quiet and 126 CBF-triggered records, and the
round-3 validation fold has 526 and 164. The guard evaluates a fixed held-out
batch $\mathcal V$ drawn from the corresponding fold rather than averaging over
the full fold. Equation~\ref{eq:guard-metrics} therefore reports full-chunk flow
loss on $\mathcal V$.

Training manifests contain 3,657, 6,535, 15,707, 19,051, and 22,732 records
over the five static rounds. Held-out records are excluded from these counts.
Quiet records constitute $56$--$62\%$ of each bank, as shown in
Figure~\ref{fig:diagnostic-panels}c.

All five adapters pass both acceptance conditions. Flow-loss ratios range from
0.9649 to 1.0363, and first-action drift ranges from 0.00514 to 0.00946. The flow-loss ratios remain below 1.10, and the first-action drifts remain below 0.05. These are checks on a mixed held-out batch, not on unseen task outcomes.
In particular, the lowest flow-loss ratio occurs at round four, when evaluation
cost reverses. Bank growth and guard acceptance alone do not establish which round best balances task success and cost.

Table~\ref{tab:bank-guard-audit} collects the review-time bank decisions. Values
not included in the audited result summaries are marked unavailable rather than
reconstructed from training artifacts. The in-loop and replacement-only banks
are not promoted because they fail at least one guard condition.

\begin{table}[htbp]
\centering
\caption{\textbf{Training-guard decisions for the bank-construction study.}
All rows use 800 update steps and one training seed. A dash denotes an
unreported diagnostic, not a zero value.}
\label{tab:bank-guard-audit}
\footnotesize
\setlength{\tabcolsep}{4.0pt}
\begin{tabular}{lrrrrl}
\toprule
Bank & Records & Flow-loss ratio & First-action drift & Triggered loss & Decision \\
\midrule
Observe-only round 1 & 3,657 & 0.972 & -- & -- & accept \\
Shield-in-loop round 1 & 5,099 & 1.137 & 0.0129 & -- & reject \\
Round 2 only & 2,878 & 1.111 & 0.0053 & -- & reject \\
Accumulated, size matched & 2,878 & 1.008 & -- & -- & accept \\
Accumulated, full & 6,535 & 1.0065 & 0.00949 & -- & accept \\
\bottomrule
\end{tabular}
\end{table}

\subsubsection{Guard-Threshold Sensitivity}
\label{app:guard-sensitivity}

The thresholds retain the implementation defaults rather than values fitted to
task SR or CC. A retrospective audit covers 113 archived validation entries,
comprising 75 metrics entries and 38 log entries. It includes repeated
representations of the same optimization run and development settings, so the
counts in Table~\ref{tab:guard-sensitivity} describe archive entries, not
independent training trials. Flow-loss ratios have median 1.000, 90th percentile
1.102, and maximum 1.580. First-action drift reaches at most 0.0160. No archived
entry exceeds the adopted drift limit of 0.05, so this limit does not affect
any observed decision.

\begin{table}[htbp]
\centering
\caption{Acceptance counts under alternative guard thresholds.}
\label{tab:guard-sensitivity}
\small
\begin{tabular}{rrrrr}
\toprule
$\tau_{\mathrm{loss}}$ & $\tau_{\mathrm{drift}}=0.01$ & 0.02 & 0.05 & 0.10 \\
\midrule
1.05 & 87 & 89 & 89 & 89 \\
1.10 & 96 & 100 & 100 & 100 \\
1.15 & 100 & 108 & 108 & 108 \\
1.20 & 100 & 109 & 109 & 109 \\
1.30 & 101 & 110 & 110 & 110 \\
\bottomrule
\end{tabular}
\end{table}

At the adopted loss limit, tightening the drift limit to 0.01 changes four
archive decisions, including the accepted $\pi_0$ quiet-regularized update.
Limits of 0.02, 0.05, and 0.10 give identical decisions in this archive, so the
flow check is the binding constraint in the observed range. Tightening the loss
limit to 1.05 flips eleven archived entries, corresponding to seven distinct
training runs. These include two main replicated updates with ratios 1.0975
and 1.0739 and the original accepted $\pi_0$ update with ratio 1.0829.
Relaxing the loss limit to 1.15 or 1.20 accepts eight or nine additional
entries, respectively. These include the in-loop and replacement-only banks,
the $\pi_0$ update with quiet-anchor weight 0.3, and the 4,800-step duration diagnostic.
Conversely, relaxing it to 1.15 admits the shield-in-loop update
whose Level 2 apple result is worse than the accepted observe-only update.
The replacement-only diagnostic is not worse despite slightly exceeding 1.10,
so these observations support an empirical, conservative guardrail rather than
an optimal or calibrated threshold. The sensitivity audit does not change the
thresholds used for the reported method.

\subsubsection{Training Duration and Guard Conservatism}
\label{app:guard-duration}

To test whether the held-out full-chunk flow-loss ratio predicts task degradation within the
main recipe, we disable the guard and vary the update duration while keeping
the bank, base checkpoint, data order, and quiet weight fixed. The cosine learning-rate schedule retains a 400-step decay period, so all subsequent updates use
the minimum learning rate of $3\times10^{-6}$. The 1,000 new evaluation cells
pass checkpoint and host checks. Each task shares its host with the base and
800-step reference. The L2 apple runs use qa-l40s-004, and the L1 tomato runs
use qa-l40s-005.

\begin{table}[htbp]
\centering
\caption{Training duration with the acceptance guard disabled.}
\label{tab:guard-duration}
\small
\setlength{\tabcolsep}{5pt}
\begin{tabular}{lrrrrrr}
\toprule
 & & & \multicolumn{2}{c}{L2 Apple} & \multicolumn{2}{c}{L1 Tomato} \\
\cmidrule(lr){4-5}\cmidrule(lr){6-7}
Steps & Flow-loss ratio & First-action drift & SR & $\mathrm{CC}_{\mathrm{policy}}$ & SR & $\mathrm{CC}_{\mathrm{policy}}$ \\
\midrule
Base & -- & -- & 8.0 & 65.89 & 85.0 & 37.01 \\
400 & 1.0042 & 0.0083 & 32.0 & 61.71 & 88.0 & 36.65 \\
800 & 1.0073 & 0.0096 & 36.0 & 65.73 & 90.0 & 31.24 \\
1,600 & 1.0165 & 0.0108 & 46.7 & 56.92 & 85.0 & 35.71 \\
2,400 & 1.0341 & 0.0121 & 54.0 & 43.01 & 90.0 & 29.38 \\
4,800 & 1.1103 & 0.0150 & 55.3 & 42.61 & 88.0 & 35.70 \\
\bottomrule
\end{tabular}
\end{table}

No tested duration produces significantly lower SR than the 800-step reference.
On L2 apple, the 2,400-step update is better under the registered paired test,
which gives a $p$-value of $0.0046$. The 400-, 1,600-, and 4,800-step comparisons are
inconclusive, with $p$-values of 0.56, 0.16, and 0.060, respectively. All SR and policy-induced CC
comparisons on L1 tomato are inconclusive. The 4,800-step adapter is the only
point exceeding the adopted loss limit of 1.10, yet its apple SR is among the
highest observed. The guard is therefore conservative over the measured flow
range of 1.004--1.110 rather than a validated predictor of task degradation.
This diagnostic does not establish behavior at ratios above 1.2. It also does
not change the main 800-step recipe. Selecting 2,400 steps after inspecting this
evaluation set would require an independent held-out test. Guard-disabled
checkpoints are diagnostic variants, not accepted main-method results.

\subsubsection{Quiet Anchors and Chunk Supervision}
\label{app:quiet-regularization}

Here $k$ denotes the number of action-chunk steps supervised during the update. The main recipe supervises the first action, and the multi-step diagnostics supervise the first 10 or all 50 actions.

The controlled $\pi_{0.5}$ quiet-anchor ablation uses the same round-2 bank,
base checkpoint, update budget, and matched L1-T3 evaluation. Changing
$\lambda_q$ from 0 to 0.2 preserves mean SR at $89.0\%$ while reducing
policy-induced CC from 20.88 to 16.52. The paired test gives a $p$-value of $0.0039$. Under this fixed recipe, quiet-anchor weighting reduces measured cost without an additional mean SR gain. The weight acts during training.

The $\pi_0$ quiet weight is selected by the held-out guard rather than by task
evaluation. Table~\ref{tab:pi0-guard-search} reports all five evaluations used
for that choice and the two multi-step follow-ups. The accepted setting has
substantially higher triggered loss than the accepted $\pi_{0.5}$ updates,
whose values range from 0.003 to 0.03.

\begin{table}[htbp]
\centering
\caption{Held-out guard search and multi-step follow-ups for $\pi_0$. The
flow-ratio limit is 1.10. Multi-step triggered losses were not reported in the
audited summary and are marked unavailable.}
\label{tab:pi0-guard-search}
\small
\begin{tabular}{rrrrl}
\toprule
Steps & $\lambda_q$ & Flow-loss ratio & Triggered loss & Decision \\
\midrule
800 & 0.0 & 1.477 & 0.185 & reject \\
500 & 0.0 & 1.513 & 0.171 & reject \\
400 & 0.0 & 1.580 & 0.170 & reject \\
800 & 0.3 & 1.110 & 0.199 & reject \\
800 & 0.5 & 1.083 & 0.178 & accept \\
800, $k=10$ & 0.0 & 1.017 & -- & accept \\
800, $k=50$ & 0.0 & 1.000 & -- & accept \\
\bottomrule
\end{tabular}
\end{table}

Passing the guard does not by itself establish a useful task update.
Table~\ref{tab:pi0-multistep} shows that supervising more of the 50-step chunk
allows the update to pass without quiet-anchor weighting, yet neither multi-step variant improves pooled SR over base. Both remain below the accepted adapter trained with first-action supervision and quiet-anchor weight 0.5.

\begin{table}[htbp]
\centering
\caption{\textbf{$\pi_0$ multi-step supervision ablation.}
Values are SR over 50 matched offsets per task. The pooled row reports mean SR
followed by mean policy-induced CC.}
\label{tab:pi0-multistep}
\footnotesize
\setlength{\tabcolsep}{6pt}
\begin{tabular}{lrrrr}
\toprule
Task & Base & $k=1,\lambda_q=0.5$ & $k=10,\lambda_q=0$ & $k=50,\lambda_q=0$ \\
\midrule
L1-T0 & 54.0 & 58.0 & 34.0 & 30.0 \\
L1-T1 & 66.0 & 84.0 & 60.0 & 62.0 \\
L1-T2 & 92.0 & 96.0 & 94.0 & 90.0 \\
L1-T3 & 28.0 & 26.0 & 24.0 & 14.0 \\
L1-T4 & 56.0 & 88.0 & 78.0 & 78.0 \\
\midrule
Pooled SR and policy-induced CC & 59.2 / 3.78 & 70.4 / 2.09 & 58.0 / 2.23 & 54.8 / 2.94 \\
\bottomrule
\end{tabular}
\end{table}

\subsubsection{Record and Teacher Threshold Sensitivity}
\label{app:label-sensitivity}

The projection trigger $z_t$ is determined by the CBF correction, not by the
distance-based risk threshold. The latter determines lead-time staging and
record admission. The safety-region detector thresholds in
Appendix~\ref{app:implementation} likewise do not define triggered or quiet
labels. We perturb the default admission risk cutoff of 0.1087 by $\pm20\%$.
The perturbed values are approximately 0.087 and 0.130. We perturb the temporal
stage boundaries of 30 and 15 steps by $\pm20\%$ and the $\eta$-bin boundaries
by $\pm0.05$.
The default reconstruction exactly reproduces the 6,535-record main bank and
its 709 genuinely corrected targets, with no record-set discrepancy.

\begin{table}[htbp]
\centering
\caption{Record and teacher threshold sensitivity.}
\label{tab:label-sensitivity}
\footnotesize
\setlength{\tabcolsep}{4pt}
\begin{tabular}{lrrrr}
\toprule
Perturbation & Bank size & Corrected targets & Affected share, \% & $\Delta$ weight mass, \% \\
\midrule
Default & 6,535 & 709 & 0.0 & 0.0 \\
Risk cutoff $\times0.8$ & 7,683 & 457 & 21.9 & $+19.3$ \\
Risk cutoff $\times1.2$ & 4,892 & 1,293 & 35.6 & $-31.0$ \\
Stage boundaries 24 / 12 & 6,626 & 744 & 1.4 & $+1.9$ \\
Stage boundaries 36 / 18 & 6,459 & 682 & 1.2 & $-1.2$ \\
$\eta$ bins $-0.05$ & 6,535 & 709 & 19.2 & $+26.3$ \\
$\eta$ bins $+0.05$ & 6,535 & 709 & 10.3 & $-14.9$ \\
\bottomrule
\end{tabular}
\end{table}

Affected percentages use the default training bank as the reference. Weight mass
is the sum of triggered-record weights. The predeclared descriptive criterion
treats at most $10\%$ affected records as
low sensitivity. Temporal boundaries meet this criterion, while risk and
$\eta$-bin thresholds do not. Changes to the $\eta$ bins preserve the record set
but alter its effective weights. Thus, the bank is not globally insensitive to
threshold choices. These are CPU-side reconstruction diagnostics without
policy retraining and do not establish corresponding SR or CC changes.

\subsubsection{Rejected-Update Diagnostics}
\label{app:rejected-diagnostics}

For diagnosis only, we retrain the two rejected bank variants after disabling
the guard while leaving all other settings unchanged. These checkpoints were rejected under the main acceptance rule and are evaluated only as diagnostic variants. Table~\ref{tab:no-guard-diag}
shows that the shield-in-loop update is worse than its accepted observe-only
counterpart on Level 2 apple, while the rejected round-2-only update does not differ significantly from the size-matched accumulated update. The latter result
shows that a guard rejection is a training-side decision rather than evidence
that accumulation improves task success.

\begin{table}[htbp]
\centering
\caption{\textbf{Non-deployable diagnostics with the acceptance guard disabled.}
Each cell reports SR followed by policy-induced CC. Bold labels mark rejected variants,
not preferred results.}
\label{tab:no-guard-diag}
\footnotesize
\setlength{\tabcolsep}{3.2pt}
\resizebox{\textwidth}{!}{%
\begin{tabular}{lrrrrrr}
\toprule
Task & Base & Observe-only R1 & \textbf{In-loop R1} & Accumulated matched & \textbf{R2 only} & Full accumulated \\
\midrule
L2-T0 Apple & 8.0 / 65.89 & 34.0 / 59.97 & 18.0 / 57.68 & 41.3 / 48.07 & 38.7 / 56.59 & 36.0 / 65.73 \\
L1-T1 Lemon & 83.0 / 35.87 & 85.0 / 24.00 & 93.0 / 3.24 & 89.0 / 10.61 & 87.0 / 8.51 & 83.0 / 12.80 \\
L1-T4 Tomato & 85.0 / 37.01 & 89.0 / 30.57 & 87.0 / 30.55 & 87.0 / 38.53 & 89.0 / 31.23 & 90.0 / 31.24 \\
\bottomrule
\end{tabular}}
\end{table}

\subsection{Diagnostics of the Dynamic-Obstacle Null Result}
\label{app:dynamic-null}

Across consecutive banks, dynamic failure profiles are at least as stable as
static ones. The mean correction-direction cosine is 0.814 for dynamic banks
and 0.747 for static banks. After controlling for episode phase, the teacher triggers more often on steps with benchmark cost than on steps without benchmark cost. The corresponding rates
are $28.6\%$ and $11.8\%$.

Correction-direction similarity and cost-step trigger rates do not support failure-mode churn or sparse triggering as explanations for the dynamic null. Triggering on cost steps does not establish that the unexecuted correction would prevent cost. They also reinforce that
intermediate quantities such as activation frequency or teacher coverage are
diagnostics rather than substitutes for matched task-level success and cost.


\section{Completion Timing and Cost Distributions}
\label{app:behavior}

\subsection{Behavior Overview}
\label{app:behavior-overview}

\begin{figure}[htbp]
\centering
\includegraphics[width=\textwidth]{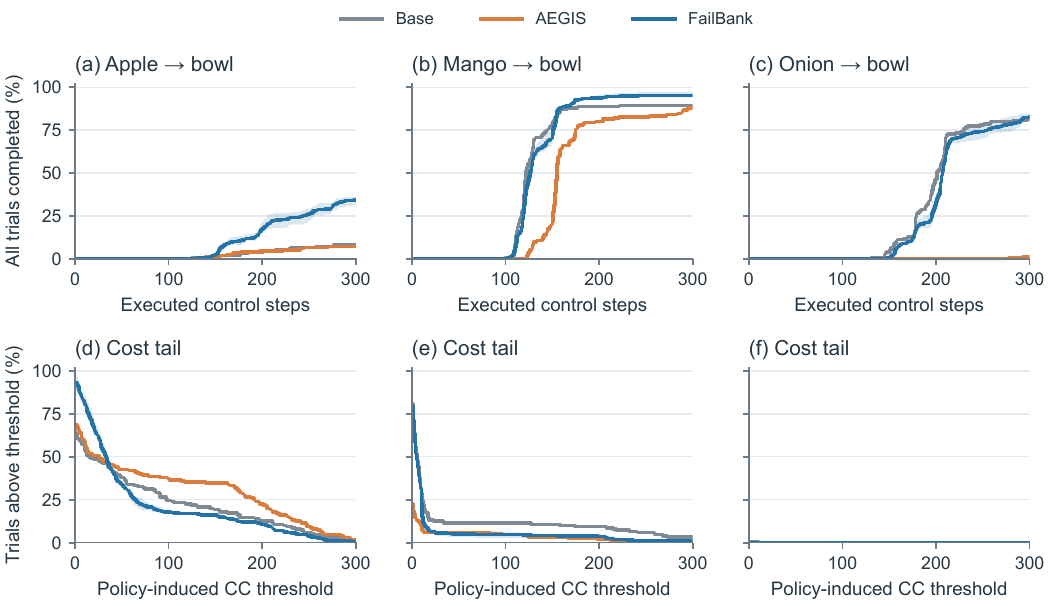}
\caption{\textbf{Completion timing and policy-induced CC distributions on Level~2 apple, mango, and onion.} Panels a--c show the fraction of all trials completed by each executed control-step count. Panels d--f show the fraction whose policy-induced CC exceeds each threshold. Blue shading spans three training orders and is not a confidence interval.}
\label{fig:behavior-panels}
\end{figure}

Figure~\ref{fig:behavior-panels} compares completion timing and policy-induced cost distributions on Level~2 apple, mango, and onion. Both analyses retain all evaluated trials, including failures. These diagnostics describe where mean SR and CC conceal differences in timing or cost distribution.

\FloatBarrier

\subsection{Policy--Shield Coordination and Collapse}
\label{sec:tradeoffs}
\label{app:coordination}

Re-attaching AEGIS tests whether the learned policy and runtime shield are
naturally compositional. If they were, the stacked system would retain the
learned success gain while lowering cost. Table~\ref{tab:stacking} instead shows
that the stacked system follows the shield's behavior on both tested tasks.

\begin{table}[htbp]
\centering
\caption{\textbf{Stacking AEGIS onto the learned policy.} Each cell reports
SR followed by $\mathrm{CC}$ over 150 matched cells.}
\label{tab:stacking}
\small
\setlength{\tabcolsep}{4.2pt}
\begin{tabular}{lrrrr}
\toprule
Task & Base & AEGIS & \method{} & \method{} + AEGIS \\
\midrule
L2-T2 Mango & 89.3 / 51.41 & 88.0 / 28.75 & 94.0 / 36.78 & 86.0 / \textbf{19.78} \\
L2-T3 Onion & 81.3 / 16.33 & 1.3 / \textbf{0.27} & \textbf{84.7} / 16.93 & 2.7 / 0.53 \\
\bottomrule
\end{tabular}
\end{table}

On mango, stacking lowers cost but decreases SR from $94.0\%$ to $86.0\%$.
On onion, stacking reduces SR from $84.7\%$ to $2.7\%$, with nearly all cells
abstaining. On both tasks, the paired test does not detect an SR difference between the stacked system and AEGIS alone. The learned behavior therefore does not preserve task progress once
the shield again controls execution.

Figure~\ref{fig:qualitative} presents one matched Level 2 onion replay. AEGIS
executes 80 CBF projections over 300 controller steps, including continuous
intervention from environment steps 40 to 88. Repeated redirection prevents a
successful grasp, and the episode times out at environment step 309. The
shield-free \method{} trajectory enters the same intervention-prone region. Its
CBF runs only in observe-only mode and flags 33 of 180 controller steps, so none of
those corrections is executed. The policy grasps and places the onion at
environment step 189 with zero policy-induced CC. Controller counts exclude
the first ten environment steps before frame logging, which explains the two
step-number conventions in the figure. This replay illustrates the mechanism,
not its frequency.

The same collapse appears without a perception module. When the
privileged CBF teacher is moved from observe-only annotation into
the execution loop, onion reaches $0\%$ SR over 150 trials. Apple falls from
$8.0\%$ to $0\%$, while mango falls from $89.3\%$ to $50.7\%$ and its
policy-induced CC rises from 33.93 to 97.50. These runs show that the loss of task completion also occurs with privileged geometry, independently of AEGIS grounding errors. Full results and configuration-validity checks appear in
Appendix~\ref{app:teacher-loop}.

\subsection{Completion Timing}
\label{app:completion-timing}

We reconstruct completion curves from the recorded success flag and executed
control-step count, including failures in the denominator. The curves therefore
do not condition on a different successful subset for each method.

By 200 steps, \method{} completes $17.3\%$ of apple trials versus $4.0\%$
for base, and $93.8\%$ of mango trials versus $88.7\%$. Onion reverses
this timing pattern. \method{} completes $33.3\%$ by that point versus
$46.7\%$ for base, despite similar final SR. The update therefore does not
uniformly speed up execution. These curves are descriptive rather than
additional significance tests at selected time thresholds.

\subsection{Policy-Induced CC Distributions}
\label{app:cost-tail}

The apple curves cross. Policy-induced CC exceeds 100 in $24.7\%$ of base
trials, $37.3\%$ with AEGIS, and $18.0\%$ with \method{}, but the fraction
with zero policy-induced CC falls from $36.0\%$ for base to $6.7\%$ for
\method{}.

Thus, the update reduces the frequency of very costly apple trials while making
nonzero cost more common. It does not dominate the distribution. On mango, the
fraction above 100 falls from $11.3\%$ to $4.7\%$. The full threshold curves
expose this dependence without promoting a selected threshold to an additional
benchmark metric.

\subsection{Completion-Coupled Cost on Onion}
\label{app:onion-cost}

CC is 16.33 for base, 0.27 for AEGIS, and 16.54 for \method{}.
The respective policy-induced CC means are 0.067, 0, and 0.004.
At least $99.3\%$ of trials in every arm have zero policy-induced CC.

Almost all of the CC difference therefore belongs to the benchmark's
initial-state/completion-coupled component. The nearly flat policy-induced cost tail shows that AEGIS sacrifices completion on a task where the logged policy-induced cost was already rare. Neither the CC gap nor the activation-rate
gap establishes a substantial contact-reduction benefit on this task.


\section{Detailed Statistical Results and Archive Eligibility}
\label{app:paired}

\subsection{Out-of-Sample Breadth and Ablation Tests}
\label{app:breadth-stats}
\label{app:full-stats}

The primary cross-task pool contains static L1-T0, T1, T3, and T4, excluding the
T2 collection task. We report the disjoint T2 state holdout separately. The
all-five-task diagnostic includes the in-sample T2 evaluation and is labeled as
such. Table~\ref{tab:breadth-paired} places paired counts outside the main
narrative, together with the learning-signal tests used in
Section~\ref{sec:learning-signals}.

\begin{table}[htbp]
\centering
\caption{\textbf{Detailed paired tests for breadth and learning-signal ablations.}
Favorable means higher SR or lower policy-induced CC. Individual L1 task rows
and legacy cost rows use the earlier no-curriculum configuration. Pooled SR rows use
training-order means.
Unreported tie counts are marked unavailable.}
\label{tab:breadth-paired}
\footnotesize
\setlength{\tabcolsep}{3pt}
\begin{tabular}{llrr}
\toprule
Analysis & Comparison & Favorable / adverse / tie & $p$-value \\
\midrule
Breadth & Static L1-T0 SR & 7 / 7 / 36 & 1.0000 \\
 & Static L1-T1 SR & 12 / 10 / 28 & 0.8320 \\
 & Static L1-T3 SR & 17 / 2 / 31 & 0.0007 \\
 & Static L1-T4 SR & 11 / 7 / 32 & 0.4810 \\
 & Four unseen L1, SR $82.5\%\to88.7\%$ & 51 / 45 / 104 & 0.61 \\
 & Four unseen tasks, legacy policy-induced CC & 71 / 53 / 76 & 0.1265 \\
 & Static L1-T2 state-holdout SR & 10 / 1 / 7 & 0.0117 \\
 & All five tasks, pooled SR & 69 / 50 / -- & 0.099 \\
 & All five tasks, legacy policy-induced CC & 101 / 70 / 79 & 0.0215 \\

\midrule
$\pi_{0.5}$ mean & Ten tasks, SR $66.0\%\to74.5\%$ & 161 / 89 / -- & $<0.0001$ \\
 & L2 five tasks, SR $50.5\%\to59.4\%$ & 92 / 39 / -- & $<0.0001$ \\
$\pi_0$ mean & Ten tasks, SR $47.5\%\to54.4\%$ & 121 / 81 / -- & 0.0059 \\
 & Four unseen L1, SR $51.0\%\to62.7\%$ & 61 / 36 / -- & 0.014 \\
 & L1 five tasks, SR $59.2\%\to68.8\%$ & 64 / 40 / -- & 0.024 \\
 & L2 five tasks, SR $35.7\%\to40.0\%$ & 57 / 41 / -- & 0.13 \\
\midrule
Learning signal & Failure-only SR relative to base & 21 / 2 / 27 & 0.0001 \\
 & Quiet-record policy-induced CC & 27 / 9 / 14 & 0.0039 \\
\bottomrule
\end{tabular}
\end{table}

AEGIS reduces Level~2 SR relative to base on both backbones. The corresponding
rates are 25.1\% versus 50.5\% on $\pi_{0.5}$ and 26.3\% versus 35.7\% on
$\pi_0$. Both paired tests give a $p$-value below $0.001$. Comparing \method{} with AEGIS over Level~2 gives
144 favorable versus 16 adverse SR directions on $\pi_{0.5}$ and 84 versus
29 on $\pi_0$. Both paired tests give a $p$-value below $0.0001$.

Across the four unseen Level~1 tasks, $\pi_{0.5}$ mean SR increases from $82.5\%$ to $88.7\%$, but its
51 wins, 45 losses, and 104 ties do not support consistent improvement across
initial states. The paired test gives a $p$-value of $0.61$. Onion carries the gain, with
$90.0\%$ SR versus $72.0\%$ and a $p$-value of $0.0026$. Collection-task mango also
improves, with a $p$-value of $0.011$. The full Level~1 mean is likewise not significant,
with a $p$-value of $0.099$. Mean Level~1 cost is descriptively lower, but the four-task
mean-based cost test is null, with a $p$-value of $0.44$.
Historical no-curriculum cost counts in Table~\ref{tab:breadth-paired} are retained
for provenance and do not establish significance for the current mean recipe.

On $\pi_0$, the reported mean uses training-order seeds 0 and 2. Seed 1 is rejected without
retraining or relaxing the guard. Its flow ratio is 1.2214 and its first-action
drift is 0.0089. Seeds 0 and 2 pass with ratios 1.0829 and 1.0171 and drifts
0.0122 and 0.0091. The Level~2 SR gain is not significant, with a $p$-value of $0.13$.
Tomato improves from $50.7\%$ to $79.3\%$, with a $p$-value of $0.0002$, but its
policy-induced CC rises from 3.61 to 6.93, with a $p$-value of $0.0026$. Seed 2 regresses on
onion from $30.7\%$ to $18.0\%$, with a $p$-value of $0.024$. Ten-task CC tests are null
on both backbones. The tests give a $p$-value of $0.085$ on $\pi_{0.5}$ and a $p$-value of $0.13$ on
$\pi_0$. The $\pi_{0.5}$ pooled Level~2 cost reduction has a $p$-value of $0.011$ and is
driven mainly by floor-task lemon. Excluding that task yields a $p$-value of $0.42$. These results do not
establish uniformly lower cost or general Level~2 transfer on $\pi_0$.

\subsubsection{Effect Sizes with Bootstrap Intervals}
\label{app:effect-sizes}

Table~\ref{tab:effect-sizes} reports learned-minus-base differences. Within
each task, we average training orders within initial state, resample initial
states 5,000 times, and average tasks equally. The primary significance test
remains the preregistered paired sign test.

\begin{table}[htbp]
\centering
\caption{\textbf{Effect sizes with 95\% bootstrap intervals.}
SR changes are percentage points. Intervals resample initial states only and
exclude variation across training orders. The non-floor L2 row is post hoc.}
\label{tab:effect-sizes}
\footnotesize
\setlength{\tabcolsep}{3pt}
\begin{tabular}{llrrrrr}
\toprule
Backbone & Set & Sign-test $p$ & $\Delta$SR & 95\% CI & $\Delta\mathrm{CC}_{\mathrm{policy}}$ & 95\% CI \\
\midrule
$\pi_{0.5}$ & L1 unseen four & 0.61 & +6.2 & [2.0, 10.5] & -8.39 & [-15.21, -1.97] \\
 & L1 all five & 0.099 & +8.3 & [4.5, 12.3] & -17.24 & [-24.22, -10.61] \\
 & L2 all five & $<0.0001$ & +8.8 & [6.2, 11.6] & -16.78 & [-24.16, -9.40] \\
 & All ten & $<0.0001$ & +8.6 & [6.2, 10.9] & -17.01 & [-22.19, -12.00] \\
$\pi_0$ & L1 unseen four & 0.014 & +11.8 & [5.0, 18.8] & -1.59 & [-3.95, 0.53] \\
 & L1 all five & 0.024 & +9.6 & [3.8, 15.6] & -1.08 & [-2.97, 0.62] \\
 & L2 all five & 0.13 & +4.3 & [0.8, 7.9] & -3.97 & [-7.71, -0.40] \\
 & L2 non-floor three & -- & +7.6 & [1.6, 13.3] & +1.82 & [1.04, 2.64] \\
 & All ten & 0.0059 & +6.9 & [3.6, 10.4] & -2.53 & [-4.63, -0.53] \\
\bottomrule
\end{tabular}
\end{table}

The positive mean-gain intervals for $\pi_{0.5}$ Level~1 across either four or five tasks
and $\pi_0$ Level~2 do not change their null sign-test conclusions. The sign test evaluates the balance of improvement and regression directions across matched initial states. The bootstrap estimates uncertainty in the mean difference. Thus, a positive mean
can coexist with inconsistent per-state improvements, as on $\pi_{0.5}$
Level~1 where onion carries much of the gain. These intervals exclude
training-order variance and can understate recipe-level uncertainty. Removing
the $\pi_0$ Level~2 floor tasks reveals increased cost, mainly on tomato.
$\pi_0$ Level~1 cost intervals include zero. Rounded task averages and
bootstrap effect sizes can differ slightly at the reported precision.

\subsection{Offset-Paired Level 2 Statistics}
\label{app:paired-stats}

Each training order is compared against the same base on 50 initial states.
For each state, we average its three sampler-advance conditions before taking
the direction of the difference. Counts are ordered as higher, lower, and equal.
They refer to the learned-minus-base direction, so higher is favorable for SR but not for
activation. All $p$-values below are two-sided exact sign-test values. Across
the 24 planned Level 2 tests, the Bonferroni-adjusted threshold is $0.00208$.
The apple SR comparisons remain supported under this threshold. The mango and
onion activation comparisons do not. A null result is not an equivalence test.

\begin{table}[htbp]
\centering
\caption{\textbf{Repeated sampler conditions are averaged, not counted as
independent samples.}
SR is success rate. Activation is the observe-only constraint-flag rate.}
\label{tab:paired-l2}
\footnotesize
\setlength{\tabcolsep}{5pt}
\begin{tabular}{lccccc}
\toprule
Task & Training &
\shortstack{SR\\higher / lower / equal} &
\shortstack{SR\\$p$-value} &
\shortstack{Activation\\higher / lower / equal} &
\shortstack{Activation\\$p$-value} \\
 & order & & & & \\
\midrule
Apple & 1 & 28 / 2 / 20 & $8.68\!\times\!10^{-7}$ & 32 / 18 / 0 & 0.0649 \\
 & 2 & 24 / 3 / 23 & $4.92\!\times\!10^{-5}$ & 22 / 26 / 2 & 0.6655 \\
 & 3 & 29 / 3 / 18 & $2.56\!\times\!10^{-6}$ & 30 / 20 / 0 & 0.2026 \\
\midrule
Mango & 1 & 8 / 4 / 38 & 0.3877 & 14 / 36 / 0 & 0.0026 \\
 & 2 & 10 / 3 / 37 & 0.0923 & 14 / 36 / 0 & 0.0026 \\
 & 3 & 10 / 7 / 33 & 0.6291 & 14 / 36 / 0 & 0.0026 \\
\midrule
Onion & 1 & 13 / 9 / 28 & 0.5235 & 34 / 16 / 0 & 0.0153 \\
 & 2 & 10 / 10 / 30 & 1.0000 & 35 / 15 / 0 & 0.0066 \\
 & 3 & 11 / 10 / 29 & 1.0000 & 28 / 22 / 0 & 0.4799 \\
\bottomrule
\end{tabular}
\end{table}

The two Level 2 tasks added for coverage were governed by a separate
pre-registered readout. Table~\ref{tab:l2-completion} gives the complete result.
Tomato satisfies the transfer criterion because two training orders improve SR
significantly and none is significantly worse than base. Lemon satisfies the
floor criterion because all methods obtain zero SR, so its cost differences are
descriptive only.

\begin{table}[htbp]
\centering
\caption{\textbf{Pre-registered Level 2 completion experiment.}
Each training-order comparison uses 50 matched offsets under three sampler
conditions. \method{} mean is the average of the three orders.}
\label{tab:l2-completion}
\footnotesize
\setlength{\tabcolsep}{4.0pt}
\begin{tabular}{llrrrr}
\toprule
Task & Arm & SR & CC & $\mathrm{CC}_{\mathrm{policy}}$ & SR test relative to base \\
\midrule
L2-T1 Lemon & Base & 0.0 & 158.14 & 158.13 & -- \\
 & AEGIS & 0.0 & 141.51 & 141.51 & -- \\
 & \method{} mean & 0.0 & 110.14 & 110.12 & floor \\
\midrule
L2-T4 Tomato & Base & 74.0 & 67.87 & 53.07 & -- \\
 & AEGIS & 28.7 & 26.93 & 21.20 & $p<0.0001$ \\
 & \method{} order 1 & 84.0 & 59.05 & 42.39 & $p=0.0872$ \\
 & \method{} order 2 & 85.3 & 55.43 & 38.49 & $p=0.0043$ \\
 & \method{} order 3 & 85.3 & 51.83 & 34.77 & $p=0.0294$ \\
 & \method{} mean & 84.9 & 55.44 & 38.55 & $p=0.0139$ \\
\bottomrule
\end{tabular}
\end{table}

Pooling all five Level 2 tasks gives mean SR $59.4\%$ for \method{} and
$50.5\%$ for base. The pooled policy-induced CC difference is driven by lemon. After
removing that floor task, policy-induced CC is 29.27 for \method{} and 38.24 for base,
and the paired difference is not significant. The paired test gives a $p$-value of $0.4159$.

Table~\ref{tab:paired-l2-cost} reports the corresponding cost comparisons.
``Favorable'' means that the learned policy has lower cost than its matched
reference. The three \method{} columns are the training orders. The AEGIS
column compares the runtime shield with its matched base.

\begin{table}[H]
\centering
\caption{Offset-paired Level 2 cost tests. $\mathrm{CC}_{\mathrm{policy}}$ removes the benchmark's
initial-state component. CC follows the benchmark definition.}
\label{tab:paired-l2-cost}
\footnotesize
\setlength{\tabcolsep}{3.5pt}
\begin{tabular}{llrrrr}
\toprule
Task & Cost & \method{}-1 & \method{}-2 & \method{}-3 & AEGIS \\
\midrule
Apple & $\mathrm{CC}_{\mathrm{policy}}$ & 25/25/0, 1.000 & 25/25/0, 1.000 & 27/23/0, 0.672 & 19/27/4, 0.302 \\
 & CC & 22/28/0, 0.480 & 23/27/0, 0.672 & 26/24/0, 0.888 & 18/27/5, 0.230 \\
\midrule
Mango & $\mathrm{CC}_{\mathrm{policy}}$ & 28/20/2, 0.312 & 34/16/0, 0.015 & 28/20/2, 0.312 & 42/8/0, $1.2{\times}10^{-6}$ \\
 & CC & 29/19/2, 0.193 & 35/15/0, 0.0066 & 29/21/0, 0.322 & 42/8/0, $1.2{\times}10^{-6}$ \\
\midrule
Onion & $\mathrm{CC}_{\mathrm{policy}}$ & 1/0/49, 1.000 & 1/1/48, 1.000 & 1/0/49, 1.000 & 1/0/49, 1.000 \\
 & CC & 9/13/28, 0.524 & 11/11/28, 1.000 & 10/11/29, 1.000 & 49/0/1, $3.6{\times}10^{-15}$ \\
\bottomrule
\end{tabular}

\vspace{2pt}
\parbox{0.98\linewidth}{\footnotesize Each cell gives favorable/adverse/tie
counts followed by the two-sided exact sign-test $p$-value. A comma separates
the counts from the $p$-value.}
\end{table}

\subsection{Archive Eligibility}
\label{app:archive}

The review batch adds 4,315 provenance-checked result cells. Every included
comparison matches the intended host and effective checkpoint. We exclude
host-inconsistent cells, replacement-only banks, obsolete correction branches,
and artifacts whose runtime audit block does not identify the intended arm.
Observe-only telemetry remains diagnostic and does not alter SR or CC. A
release archive should include per-cell result files, adapter audit blocks,
training and bank manifests, and recovery logs needed to reconstruct each
reported aggregate.

\end{document}

%% file: math_commands.tex
\usepackage{amsmath,amsfonts,bm}

\def\eqref#1{equation~\ref{#1}}

\def\1{\bm{1}}

\DeclareMathAlphabet{\mathsfit}{\encodingdefault}{\sfdefault}{m}{sl}
\SetMathAlphabet{\mathsfit}{bold}{\encodingdefault}{\sfdefault}{bx}{n}



%% file: figures/summary_table_two_backbones.tex
\begin{table}[t]
\centering
\caption{\textbf{Static-obstacle results across two backbones.} Collection happens in Level~1 Mango task while other tasks are unseen during rollout collection.
SR is task success rate (\%). CC is official cumulative cost.
$\mathrm{CC}_{\mathrm{policy}}$ is policy-induced cost. Best SR, $\mathrm{CC}_{\mathrm{policy}}$ and BRS
values are in \textbf{bold}.}
\label{tab:summary}

\setlength{\tabcolsep}{2.4pt}
\renewcommand{\arraystretch}{1.08}
\resizebox{\textwidth}{!}{%
\begin{tabular}{ll|ccc|ccc|ccc||ccc|ccc|ccc}
\toprule
 & & \multicolumn{9}{c||}{$\pi_{0.5}$} & \multicolumn{9}{c}{$\pi_{0}$} \\
\cmidrule(lr){3-11}\cmidrule(lr){12-20}
 & & \multicolumn{3}{c|}{Base} & \multicolumn{3}{c|}{AEGIS} & \multicolumn{3}{c||}{\method{}}
   & \multicolumn{3}{c|}{Base} & \multicolumn{3}{c|}{AEGIS} & \multicolumn{3}{c}{\method{}} \\
Level & Task & SR$\uparrow$ & CC$\downarrow$ & $\mathrm{CC}_{\mathrm{policy}}\downarrow$
 & SR$\uparrow$ & CC$\downarrow$ & $\mathrm{CC}_{\mathrm{policy}}\downarrow$
 & SR$\uparrow$ & CC$\downarrow$ & $\mathrm{CC}_{\mathrm{policy}}\downarrow$
 & SR$\uparrow$ & CC$\downarrow$ & $\mathrm{CC}_{\mathrm{policy}}\downarrow$
 & SR$\uparrow$ & CC$\downarrow$ & $\mathrm{CC}_{\mathrm{policy}}\downarrow$
 & SR$\uparrow$ & CC$\downarrow$ & $\mathrm{CC}_{\mathrm{policy}}\downarrow$ \\
\midrule
\multirow{5}{*}{1}
 & Apple  & 90.0 & 9.12 & 0.12 & 82.0 & 8.20 & \textbf{0.00} & \textbf{90.3} & 9.21 & 0.21
          & 54.0 & 5.40 & \textbf{0.00} & \textbf{82.0} & 8.20 & \textbf{0.00} & 59.0 & 5.90 & \textbf{0.00} \\
 & Lemon  & \textbf{83.0} & 44.07 & 35.87 & 78.0 & 7.81 & \textbf{0.01} & 81.0 & 25.98 & 17.88
          & 66.0 & 6.60 & \textbf{0.00} & 80.0 & 8.00 & \textbf{0.00} & \textbf{84.0} & 8.40 & \textbf{0.00} \\
 & Mango  & 77.0 & 74.08 & 66.38 & 68.0 & 21.08 & 14.28 & \textbf{93.7} & 23.12 & \textbf{13.76}
          & 92.0 & 11.34 & 2.14 & 82.0 & 9.40 & \textbf{1.20} & \textbf{93.0} & 12.39 & 3.09 \\
 & Onion  & 72.0 & 33.88 & 27.18 & 44.0 & 9.12 & \textbf{4.71} & \textbf{90.0} & 31.63 & 23.50
          & 28.0 & 18.34 & 16.74 & \textbf{30.0} & 3.00 & \textbf{0.00} & 24.0 & 12.02 & 10.32 \\
 & Tomato & 85.0 & 45.31 & 37.01 & \textbf{94.0} & 17.39 & \textbf{7.99} & 93.3 & 34.28 & 25.04
          & 56.0 & 5.60 & \textbf{0.00} & 44.0 & 4.40 & \textbf{0.00} & \textbf{84.0} & 8.46 & 0.06 \\
\midrule
\multirow{5}{*}{2}
 & Apple  & 8.0 & 67.51 & 65.89 & 7.3 & 90.73 & 89.26 & \textbf{34.2} & 68.75 & \textbf{61.90}
          & 0.0 & 66.31 & 66.31 & 0.0 & 54.63 & 54.62 & \textbf{0.3} & 50.29 & \textbf{50.22} \\
 & Lemon  & \textbf{0.0} & 158.14 & 158.13 & \textbf{0.0} & 141.51 & 141.51 & \textbf{0.0} & 110.14 & \textbf{110.12}
          & 2.0 & 14.09 & 13.69 & \textbf{10.7} & 7.39 & 5.26 & 0.3 & 4.53 & \textbf{4.46} \\
 & Mango  & 89.3 & 51.41 & 33.93 & 88.0 & 28.75 & \textbf{11.15} & \textbf{95.1} & 35.24 & 16.61
          & \textbf{95.3} & 22.47 & 3.67 & 94.7 & 19.61 & \textbf{0.67} & 94.7 & 24.41 & 5.80 \\
 & Onion  & 81.3 & 16.33 & 0.067 & 1.3 & 0.27 & \textbf{0.000} & \textbf{82.7} & 16.54 & 0.004
          & \textbf{30.7} & 6.13 & \textbf{0.000} & 6.7 & 1.34 & 0.007 & 25.3 & 5.07 & \textbf{0.000} \\
 & Tomato & 74.0 & 67.87 & 53.07 & 28.7 & 26.93 & \textbf{21.20} & \textbf{84.9} & 55.44 & 38.55
          & 50.7 & 13.75 & 3.61 & 19.3 & 4.81 & \textbf{0.95} & \textbf{79.3} & 22.67 & 6.93 \\
\midrule
\multicolumn{2}{l|}{BRS$\uparrow$}
 & \multicolumn{3}{c|}{0.368} & \multicolumn{3}{c|}{0.350} & \multicolumn{3}{c||}{\textbf{0.498}}
 & \multicolumn{3}{c|}{0.368} & \multicolumn{3}{c|}{0.441} & \multicolumn{3}{c}{\textbf{0.443}} \\
\bottomrule
\end{tabular}}
\end{table}